\documentclass[lettersize,journal]{IEEEtran}

\usepackage{amsmath,amsfonts}
\usepackage{array}
\usepackage{textcomp}
\usepackage{stfloats}
\usepackage{float}
\usepackage{caption}
\usepackage{url}
\usepackage{verbatim}
\usepackage{graphicx}
\usepackage{cite}
\usepackage{multicol}
\usepackage{newtxtext}
\usepackage{newtxmath}
\usepackage{xcolor}
\usepackage{svg}
\usepackage{booktabs}
\usepackage{multirow}
\usepackage{makecell}
\usepackage[normalem]{ulem}

\DeclareMathOperator*{\argmax}{arg\,max}

\usepackage{algorithm}
\usepackage{algpseudocode}
\algnewcommand\algorithmicforeach{\textbf{for each}}
\algdef{S}[FOR]{ForEach}[1]{\algorithmicforeach\  #1\ \algorithmicdo}
\usepackage{caption}

\usepackage{subcaption}
\usepackage{placeins}

\newcommand*{\algo}{VQ-Elites}
\newcommand*{\aur}{AURORA$^\dagger$}
\newcommand{\change}[1]{#1}

\usepackage{verbatim}
\def\BibTeX{{\rm B\kern-.05em{\sc i\kern-.025em b}\kern-.08em
    T\kern-.1667em\lower.7ex\hbox{E}\kern-.125emX}}
\usepackage{balance}

\usepackage{url}
\usepackage{hyperref}
\usepackage{eso-pic}

\begin{document}
\AddToShipoutPicture*{%
    \AtPageLowerLeft{%
        \parbox[t][\paperheight][t]{\paperwidth}{%
            \vspace*{-1.5cm}
            \begin{center}

            \textbf{Full Reference:} C. Tsakonas et al. ``Vector Quantized-Elites: Unsupervised and Problem-Agnostic Quality-Diversity Optimization''\\ in IEEE Transactions on Evolutionary Computation, Nov. 2025, \\
            \href{https://doi.org/10.1109/TEVC.2025.3631786}{https://doi.org/10.1109/TEVC.2025.3631786}
        \end{center}
        }
    }
}

\title{Vector Quantized-Elites: Unsupervised and Problem-Agnostic Quality-Diversity Optimization}

\author{Constantinos Tsakonas$^{1\dagger}$ and Konstantinos Chatzilygeroudis$^{1,2}$
\thanks{$^{1}$Computational Intelligence Laboratory (CILab), Department of Mathematics,
        University of Patras, GR-26110 Patras, Greece}%
\thanks{$^{2}$Laboratory of Automation and Robotics (LAR) in the Department of Electrical \& Computer Engineering,
        University of Patras, GR-26504 Patras, Greece,
        {\tt\small costashatz@upatras.gr}}%
\thanks{$^{\dagger}$C. Tsakonas is now with Inria Centre at Université de Lorraine - work done while at CILab, {\tt\small konstantinos.tsakonas@inria.fr}}%
}

\markboth{IEEE Transactions on Evolutionary Computation}{Tsakonas and Chatzilygeroudis: Vector Quantized-Elites: Unsupervised and Problem-Agnostic Quality-Diversity Optimization}

\maketitle

\begin{abstract}
Quality-Diversity algorithms have transformed optimization by prioritizing the discovery of diverse, high-performing solutions over a single optimal result. However, traditional Quality-Diversity methods, such as MAP-Elites, rely heavily on predefined behavior descriptors and complete prior knowledge of the task to define the behavior space grid, limiting their flexibility and applicability. In this work, we introduce Vector Quantized-Elites (VQ-Elites), a novel Quality-Diversity algorithm that autonomously constructs a structured behavior space grid using unsupervised learning, eliminating the need for prior task-specific knowledge. At the core of VQ-Elites is the integration of Vector Quantized Variational Autoencoders, which enables the dynamic learning of behavior descriptors and the generation of a structured, rather than unstructured, behavior space grid --- a significant advancement over existing unsupervised Quality-Diversity approaches. This design establishes VQ-Elites as a flexible, robust, and task-agnostic optimization framework. To further enhance the performance of unsupervised Quality-Diversity algorithms, we introduce behavior space bounding and cooperation mechanisms, which significantly improve convergence and performance, as well as the Effective Diversity Ratio and Coverage Diversity Score, two novel metrics that quantify the actual diversity in the unsupervised setting. We validate VQ-Elites on robotic arm pose-reaching, mobile robot space-covering, and MiniGrid exploration tasks. The results demonstrate its ability to efficiently generate diverse, high-quality solutions, emphasizing its adaptability, scalability, robustness to hyperparameters, and potential to extend Quality-Diversity optimization to complex, previously inaccessible domains.
\end{abstract}


\section{Introduction}

\begin{figure*}[t!]
      \includegraphics[width=\linewidth]{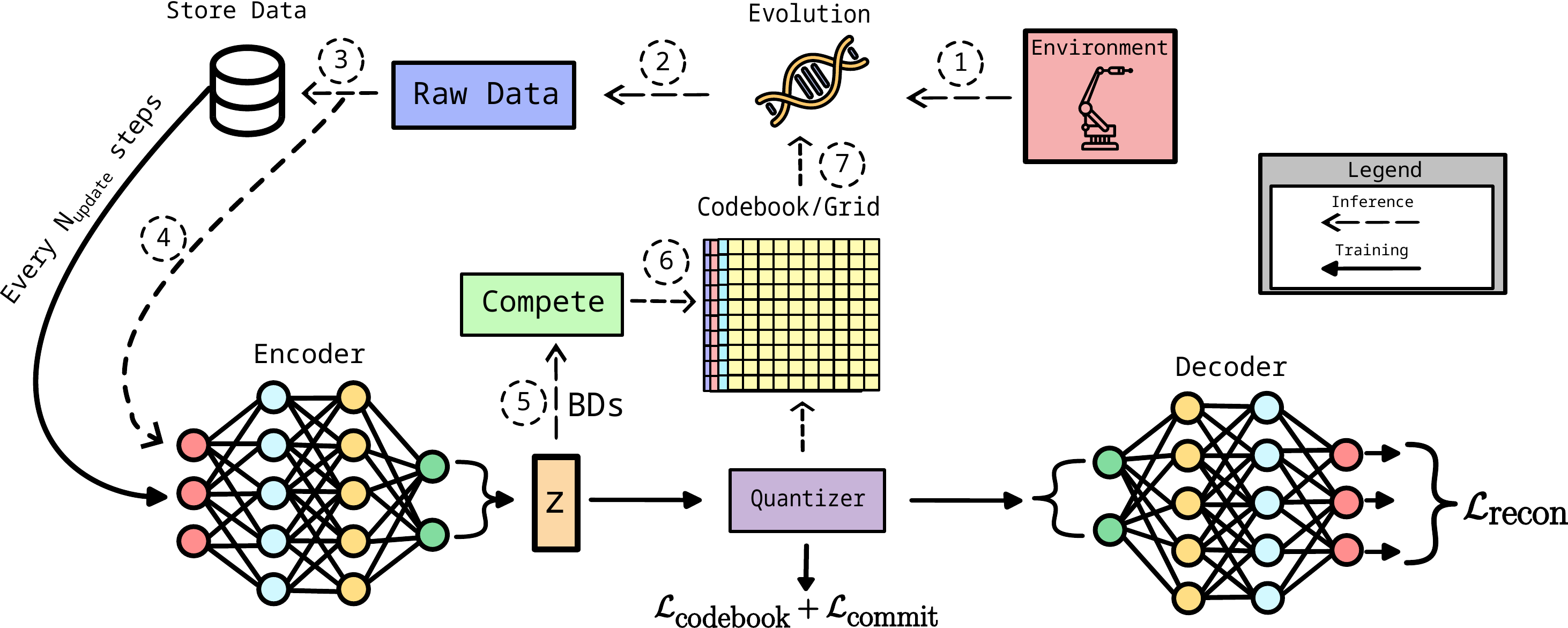}
      \caption{Overview of \algo{}. The algorithm and each component depicted in the figure are described in detail in Sec.~\ref{sec:methodology}.}
      \label{fig:overview}
\end{figure*}

\IEEEPARstart{T}{raditional} evolutionary algorithms (EAs) have been a cornerstone of optimization, designed primarily to identify a single optimal solution for a given problem. Although effective for well-defined tasks, this narrow focus on singular objectives limits their ability to address complex challenges, such as those in robotics and autonomous systems. This limitation hinders their adaptability in dynamically changing environments, where a diverse range of high-performing solutions is often required. Consequently, traditional EAs struggle in scenarios that demand versatility rather than a single, best solution~\cite{cully2017quality,chatzilygeroudis2021quality}.

Quality-Diversity (QD) algorithms address the limitations of traditional evolutionary algorithms by redefining the optimization objective~\cite{cully2017quality,chatzilygeroudis2021quality}. Rather than seeking a single optimal solution, QD algorithms, such as MAP-Elites~\cite{mouret2015illuminatingsearchspacesmapping,cully2015robots}, aim to uncover a diverse set of high-performing solutions across a broad behavior landscape. This ``illumination'' approach generates multiple effective individuals, each associated with a unique behavior. By promoting both diversity and quality, QD algorithms enable more flexible and robust applications, particularly in autonomous systems, where a repertoire of high-quality behaviors is essential for adapting to dynamic challenges. For instance, QD algorithms have demonstrated the ability to generate diverse locomotion behaviors in quadruped robots~\cite{cully2015robots,chatzilygeroudis2018reset,loco_srbd}.

Despite these advances, QD algorithms face significant limitations, particularly regarding the reliance on predefined or "hard-coded" behavior descriptors (BDs). In MAP-Elites, for example, the behavior space is often discretized using a set of user-defined descriptors to categorize behaviors, which imposes constraints on the algorithm's adaptability. These hard-coded descriptors assume a priori knowledge about the task or environment and may not fully capture the complexity or diversity inherent in high-dimensional, novel tasks. This constraint limits the algorithm’s flexibility, as predefined descriptors may inadvertently filter out behaviors that fall outside of the original descriptors or misrepresent complex solutions. Moreover, for tasks where the behavior descriptors are hard to design or even unknown, the manual design of BDs becomes impractical and can hinder the QD algorithm's capacity to scale or generalize to new domains.

Early approaches to automatic BD extraction relied on large static datasets to learn lower-dimensional representations using techniques like Principal Component Analysis (PCA) or auto-encoders. For instance, Cully and Demiris~\cite{cully2018hierarchical} utilized a dataset of handwritten digits to train an auto-encoder, creating a low-dimensional latent space that captured the key features of different digits. This latent space then served as the behavior descriptor space, enabling QD algorithms to generate a diverse set of solutions spanning the learned feature space. This approach allowed a robotic arm to autonomously learn to draw digits without requiring manual definitions of their key characteristics.

The challenges posed by predefined BDs and static datasets encouraged the development of unsupervised learning methods for BD extraction. A notable example is the AURORA algorithm~\cite{cully2019autonomous,aurora}, which employs neural network-based representation learning to autonomously identify behavior descriptors directly from the solution space. The main innovation compared to using a static dataset is that AURORA uses the data generated by the QD algorithm to train the auto-encoder in an online fashion. This innovation represented a significant leap, allowing QD algorithms to generalize across a broader range of tasks without relying on hand-crafted descriptors, thus enabling more adaptable and versatile QD frameworks. Nevertheless, AURORA still requires a lot of heuristics and extensive fine-tuning to be performant.

In this paper, we introduce Vector Quantized-Elites (\algo), a novel QD algorithm that uses unsupervised learning to autonomously construct meaningful behavior spaces without prior task or environment knowledge (Fig.~\ref{fig:overview}). We leverage a Vector Quantized Variational Autoencoder (VQ-VAE) to make our algorithm able to dynamically learn behavior descriptors, thus enabling autonomous discovery of relevant skills and behaviors. By integrating VQ-VAE into the QD framework, our approach organizes the behavior space into a structured grid by automatically clustering latent representations. This grid-based structure addresses the shortcomings of previous unsupervised BD methods by combining flexibility with the advantages of well-structured archives. Moreover, VQ-Elites enhances the scalability of QD algorithms, facilitating exploration in high-dimensional tasks and environments that traditional QD methods struggle to handle.

To demonstrate the effectiveness of VQ-Elites, we evaluate its performance across a variety of tasks that highlight its adaptability and robustness in complex scenarios. These include both structured and unstructured space-covering tasks, a hard exploration task, as well as high-dimensional pose-reaching tasks for robotic manipulators. In each experiment, we assess VQ-Elites’ ability to generate diverse, high-quality behaviors while maintaining an organized archive. The results underscore its strengths in flexibility and robustness, showcasing its potential to advance QD frameworks
and extend their applicability to more complex problem domains.

In short:
\begin{itemize}
    \item We propose \textbf{\algo}, a novel QD algorithm that integrates unsupervised learning to construct a flexible, task-agnostic behavior space. Using VQ-VAE models, \algo~dynamically learns BDs, enabling the autonomous discovery of relevant behaviors across diverse tasks without prior environment-specific knowledge.
    \item \algo~\emph{autonomously} creates a \textbf{structured grid in behavior space} by clustering latent representations, addressing the limitations of previous unsupervised BD methods while maintaining the benefits of structured archives.
    \item We introduce two components, \textbf{Behavior Space Bounding} and \textbf{Cooperation}, which drastically improve the convergence of unsupervised QD algorithms.
    \item We propose two novel metrics, \textbf{Effective Diversity Ratio} and \textbf{Coverage Diversity Score}, to properly quantify the performance of unsupervised QD approaches.
    \item We evaluate \algo~across diverse tasks, including space-covering and hard exploration challenges, pose-reaching tasks for robotic manipulators, and scenarios involving high-dimensional data, and constrained systems.
    \item Experimental results show that \algo~efficiently generates diverse, high-quality behaviors while preserving a well-organized archive, highlighting its potential to enhance QD frameworks
    and extend their applicability to complex problem domains.
\end{itemize}
\section{Preliminaries}
\subsection{Quality-Diversity Algorithms}

One of the notable strengths of QD algorithms is their ability to provide multiple high-performing solutions, offering users significant flexibility~\cite{cully2017quality,chatzilygeroudis2021quality}. This flexibility arises from the wide range of solutions available, allowing users to balance complexity, desired behavior, and practical implementation. Here, we present shortly the QD optimization framework for completeness, but we refer the reader to recent surveys for a more detailed overview~\cite{chatzilygeroudis2021quality,cully2017quality}.

To express this more formally, consider an objective function that returns a fitness value, $f_\theta$, and a behavior descriptor vector, $\boldsymbol{b}_\theta$, which defines the behavior:
\begin{align}
f_{\theta}, \boldsymbol{b}_{\theta} \leftarrow f(\boldsymbol{\theta})
\end{align}

In practice, the behavior descriptor characterizes the solution, while the fitness value measures its quality. To clarify the concept of a BD and the difference to the fitness, consider a scenario where a robot must traverse various locations within an environment. In this case, the robot's coordinates can be the BD, while the average speed it reaches the targets the fitness value. Bringing these components together, the optimization algorithm generates multiple solutions, each defined by its BD vector, which describes the behavior or outcome, and its fitness value, which quantifies its effectiveness.

\textit{But, what is our ultimate goal when using a Quality-Diversity algorithm?} Let us assume, without the loss of generality, that we have a behavior space (feature space) $\mathcal{B}$ and a fitness function that we want to maximize. The ultimate goal of QD optimization is to find all individuals\footnote{In our experiments, we always use real-valued parameter vectors.}, $\boldsymbol{\theta}$, that maximize the fitness function for every $\boldsymbol{b}\in\mathcal{B}$:
\begin{align}
    \forall \boldsymbol{b} \in \mathcal{B} \qquad \boldsymbol{\theta}^{*} = \argmax_{\boldsymbol{\theta}}f_{\theta} \nonumber \\
    s.t. \qquad \boldsymbol{b} = \boldsymbol{b_{\theta}}
\end{align}

As already mentioned, the result of a QD algorithm is a set of solutions. This set is called either the \textit{archive}, \textit{collection}, or \textit{map}. Furthermore, during the optimization process, it is possible that the algorithm can produce a solution with a similar BD to a solution that already exists in the archive. In that case, and since we cannot keep all possible solutions, the two solutions compete, and finally, the fittest one is preserved.

Archives in QD algorithms can take various forms. Perhaps the easiest one involves discretizing the behavior space into a grid, where each cell corresponds to a unique solution. Solutions generated by the algorithm are assigned to the closest cell based on the Euclidean distance between the solution's BD and the cell's center. This grid-based method is widely used, particularly in the MAP-Elites algorithm, one of the most popular QD optimization approaches. Other archive variations include more abstract and unstructured designs. For instance, a user may define a distance threshold, allowing a solution to be stored if its distance to all existing BDs exceeds the threshold, or if it outperforms competing BDs within the threshold.

The MAP-Elites algorithm~\cite{mouret2015illuminatingsearchspacesmapping} is one of the first and most popular QD algorithms.
The popularity of MAP-Elites stems from its simplicity, both in terms of understanding and implementation. However, its main challenge lies in discretizing the behavior space, which is highly problem- and domain-dependent. This becomes particularly difficult in scenarios involving high-dimensional features or when the behavior space lacks well-defined constraints.

\subsection{Unsupervised Representation Learning}
Representation learning is a fundamental concept in machine learning and artificial intelligence, focusing on the development of effective data representations that enhance the performance of machine learning models on various tasks. The goal is to automatically transform raw data, such as images, text, or audio, into a more abstract and useful format that captures the essential features of the data while discarding irrelevant or redundant information. Effective representations can simplify learning tasks, reduce the need for domain-specific feature engineering, and enable generalization across different datasets and tasks. In contrast, traditional approaches relied heavily on manual feature engineering, requiring domain experts to identify and extract useful features. This process, however, is often labor-intensive, prone to biases, and limits scalability across different problem domains. Representation learning automates this process, allowing models to discover and learn optimal features from data autonomously. 

VQ-VAE represents a powerful extension of traditional autoencoders, combining ideas from variational autoencoders (VAEs) and vector quantization. Introduced by Oord et al.~\cite{vqvae_paper}, VQ-VAE addresses the limitations of continuous latent spaces by discretizing them, allowing the model to learn discrete latent representations that are often more interpretable and suitable for tasks such as generation, compression, and unsupervised learning.

VQ-VAE introduces a discrete latent space by using \textit{vector quantization}. The encoder maps the input $\boldsymbol{x}$ to a latent representation, but instead of sampling from a continuous distribution, the latent code is quantized by selecting the closest code from a predefined set of embeddings, known as the \textit{codebook}. Let $\mathcal{C} \in \mathbb{R}^{K \times D}$ represent the codebook, where $K$ is the number of embeddings and $D$ is the dimensionality of each embedding vector. For each input $\boldsymbol{x}$, the encoder produces a latent vector $\boldsymbol{z}_e(\boldsymbol{x})$, and the quantized vector $\boldsymbol{z}_q(\boldsymbol{x})$ is obtained by choosing the nearest codebook entry using the following operation:

\begin{equation}
\boldsymbol{z}_q(\boldsymbol{x}) = \arg \min_{\boldsymbol{c}_i \in \mathcal{C}} \| \boldsymbol{z}_e(\boldsymbol{x}) - \boldsymbol{c}_i \|^2
\end{equation}

This step discretizes the latent space, as the latent vector $\boldsymbol{z}_q(\boldsymbol{x})$ is always chosen from a finite set of possible embeddings $\{\boldsymbol{c}_1, \boldsymbol{c}_2, \dots, \boldsymbol{c}_K\}$.
The training of a VQ-VAE is performed through back-propagation with the goal of minimizing a loss function. The overall loss function of VQ-VAE is composed of three key components:

\begin{itemize}
    \item \textbf{Reconstruction Loss:} This term ensures that the decoder reconstructs the input data accurately from the quantized latent code $\boldsymbol{z}_q(\boldsymbol{x})$. Similar to traditional autoencoders, the reconstruction loss is typically measured as the $L_2$-norm between the input and the reconstructed output $\hat{\boldsymbol{x}}$:
    \begin{equation}
    \mathcal{L}_{\text{recon}} = \| \boldsymbol{x} - \hat{\boldsymbol{x}} \|^2
    \end{equation}

    \item \textbf{Codebook Loss (Commitment Loss):} Since the latent space is quantized, the encoder is encouraged to produce latent vectors that are close to the nearest codebook vectors. This is done by adding a commitment loss that penalizes large differences between $\boldsymbol{z}_e(\boldsymbol{x})$ and $\boldsymbol{z}_q(\boldsymbol{x})$. The commitment loss pushes the codebook vectors to move towards the encoded latent vectors, ensuring that the quantized vectors are close representations of the original latent vectors:
    \begin{equation}
    \mathcal{L}_{\text{codebook}} = \| \texttt{sg}[\boldsymbol{z}_e(\boldsymbol{x})] - \boldsymbol{c} \|^2
    \end{equation}
    where $\texttt{sg}[\cdot]$ represents the stop-gradient operator, which prevents gradients from being back-propagated through the codebook vectors.

    \item \textbf{Commitment Loss for Encoder:} To prevent the encoder from fluctuating too far from the chosen codebook vectors, a commitment loss is also applied to the encoder:
    \begin{equation}
    \mathcal{L}_{\text{commit}} = \beta \| \boldsymbol{z}_e(\boldsymbol{x}) - \texttt{sg}[\boldsymbol{c}] \|^2
    \end{equation}
    Here, $\beta$ is a hyperparameter that controls the trade-off between minimizing reconstruction error and ensuring the encoder’s output aligns with the quantized latent space.
\end{itemize}

Thus, the total VQ-VAE loss is given by:

\begin{equation}
\mathcal{L}_{\text{VQ-VAE}} = \mathcal{L}_{\text{recon}} + \mathcal{L}_{\text{codebook}} + \mathcal{L}_{\text{commit}}
\end{equation}

Essentially, despite the reconstruction loss that is present in all the encoder-decoder type of architectures, the two new losses operate in a complementary way. The codebook loss adapts the dictionary, while the commitment loss regularizes the encoder. Thus, the former encourages the codebook embeddings to move closer to the encoder outputs, effectively updating the codebook to represent the data more faithfully. On the other hand, the latter acts in the reverse direction, penalizing the encoder for straying too far from the chosen codebook vectors, therefore ensuring stability and consistency in encoding.

One of the key challenges of minimizing this combined loss is the non-differentiability of the quantization step (i.e., selecting the nearest codebook entry), which prevents gradients from flowing back through the encoder during standard training. To address this, VQ-VAE uses a \textit{straight-through estimator} for gradient updates. Specifically, during the backward pass, the gradients from the decoder are passed through the quantization operation unchanged, which allows effective training despite the discrete nature of the latent space~\cite{vqvae_paper}.
This paradigm shift to discrete latent spaces has led to breakthroughs in various fields, including natural language processing~\cite{DBLP:journals/corr/abs-2312-11532}, computer vision~\cite{vq-vae-2, walker2021predictingvideovqvae}, speech recognition~\cite{DBLP:conf/icassp/WilliamsZCY21, DBLP:conf/icassp/ZhouBA21}, and reinforcement learning~\cite{agrl, edl}.
\section{Related Work}

The MAP-Elites algorithm, introduced by Mouret and Clune~\cite{mouret2015illuminatingsearchspacesmapping}, established a new paradigm in evolutionary optimization by illuminating diverse regions of the solution space rather than focusing solely on a single optimal solution. This class of algorithms later became known as Quality-Diversity algorithms~\cite{pugh2016quality}. MAP-Elites partitions the feature space into user-defined niches and searches for the highest-performing solution within each niche. This approach has been utilized in multiple domains, showcasing its applicability and impact, regarding the robustness of LLMs in adversarial prompts~\cite{rainbow_teaming}, game generation~\cite{gavel}, evolving graph structures~\cite{nadizar2024searching}, image synthesis~\cite{me_image_synthesis}. Additionally, MAP-Elites has been highly influential in robotics, specifically in robotic control tasks and constrained optimization, which require diverse behaviors (features)~\cite{cully2015robots}. However, MAP-Elites relies heavily on pre-defined behavior descriptors (BDs), which limits its flexibility and adaptability to dynamic or complex environments where optimal descriptors are not readily apparent.

To address this limitation, various adaptations have been proposed. The CVT-MAP-Elites algorithm uses centroidal Voronoi tessellations (CVTs) to partition the behavior space based on well-spread centroids, allowing for more natural, high-dimensional behavior mappings without rigid grid structures~\cite{map-elites-voronoi}. Despite its effectiveness in certain applications, CVT-MAP-Elites still depends on predefined centroids, which may not capture the full range of possible behaviors in a dynamic environment.

More recent work has explored integrating reinforcement learning (RL) techniques with MAP-Elites to improve search efficiency and expand behavior diversity. For example, methods like PGA-MAP-Elites and QD-PG leverage policy gradients alongside MAP-Elites’ mutation-based search, enabling gradient-informed exploration of high-dimensional behaviors~\cite{nilsson2021policy, pierrot2022qd}. However, these approaches still rely on predefined descriptors or additional complexity from combining RL methods, which may not be ideal for fully unsupervised BD discovery.

In parallel, advancements in skill discovery through hierarchical reinforcement learning (HRL) and skill chaining~\cite{barto2003recent, konidaris2009skill} have shown the importance of decomposing tasks into reusable sub-skills. These methods provide a framework for discovering hierarchical behaviors but often require substantial supervision or domain knowledge. Unsupervised skill discovery methods, such as DIAYN (Diversity is All You Need)~\cite{eysenbach2018diversity} and VALOR~\cite{gregor2016variational}, emphasize maximizing diversity in the skill space while learning useful policies. However, these approaches primarily focus on policy-level diversity without leveraging structured archives like MAP-Elites for comprehensive exploration.

The introduction of unsupervised learning methods into the QD framework, like AURORA~\cite{aurora}, has enabled the automatic extraction of latent behavior representations, allowing QD algorithms to operate independently of pre-defined descriptors. For instance, AURORA employs neural networks to identify meaningful patterns within the behavior space, yet it lacks a structured archive organization akin to MAP-Elites, which can hinder systematic exploration in complex domains~\cite{aurora} (as we will show in our experiments). The TAXONS algorithm~\cite{9196819}, the predecessor of AURORA, also utilizes an autoencoder with RGB images as input to construct a low-dimensional behavior space, or outcome space as they characterize it. The authors, also use two metrics to decide the fitness of a solution, the fitness score and the autoencoder's reconstruction error, and one of them is chosen for each iteration with a probability of 50\%. More recently, STAX~\cite{paolo2024discovering} operates through an alternating two-phase process: exploration, where it builds a diverse set of policies and learns the behavior representation; and exploitation, where it optimizes the performance of policies that have discovered rewards. Similarly, techniques leveraging deep reinforcement learning~\cite{ecoffet2021go} and self-supervised approaches~\cite{gupta2018unsupervised} have demonstrated promise in autonomous skill discovery, but their reliance on task-specific tuning can restrict generalization. Lastly, the Quality-Diversity Meta-Evolution (QD-Meta) framework~\cite{9716036}, which evolves a population of QD algorithms to dynamically customize behavior spaces for user-defined meta-objectives. By using the Covariance Matrix Adaptation Evolutionary Strategy (CMA-ES), QD-Meta uses neural networks that transform high-dimensional behavior features into compact descriptors while leveraging reinforcement learning for dynamic parameter control. This approach enables the creation of tailored, diverse, and high-quality solutions across various domains without relying on manually crafted behavior spaces. Our approach, \algo, builds on these advancements by combining MAP-Elites with a VQ-VAE to autonomously generate structured and flexible behavior spaces, thereby overcoming the limitations of previous unsupervised methods.
\section{Vector-Quantized Elites} \label{sec:methodology}
The primary goal of this work is to advance the MAP-Elites to operate entirely unsupervised, eliminating the need for prior knowledge about the task, environment, or dynamic constraints, and in the process to propose a new paradigm for the development of unsupervised QD algorithms. Our proposed \algo~algorithm achieves this by leveraging the agent's experiences in conjunction with Vector Quantized machine learning models. This combination allows \algo~to dynamically and autonomously
adapt to the specific characteristics of the problem at hand, ensuring flexibility and scalability across diverse tasks and environments. A detailed overview of the algorithm is provided in Algo.~\ref{algo:vq-eltes} and in Fig.~\ref{fig:overview}.
\begin{algorithm*}
    \caption{\algo~algorithm (number of iterations $N_\text{evo}$; model update interval $N_\text{update}$; training epochs $\mathcal{E}$; cooperation iterations $N_\text{cooperation}$)}\label{algo:vq-eltes}
    \begin{algorithmic}[1]
    
        \State \textbf{Define:} $\mathcal{M},\; \mathcal{C}$ \Comment{VQ-VAE model, behavior space grid}
        \State \textbf{Initialize:} $\mathcal{D} \leftarrow \varnothing$, $\mathcal{M}_\text{codebook}$ \Comment{Initialize an empty dataset, Initialize model's codebook randomly or using K-Means}
        
        \Procedure{VQ-Elites}{}
        \State $\mathcal{A} \longleftarrow \{\varnothing\}_{|C|}$ 
        \Comment{Initialize an empty archive of size $|C|$ (number of cells in the grid)}
        
        \State \textsc{Bootstrap()}
        
        \For{$i = 1 \to I$}
            \State \textsc{Evolution Phase()}
            \Comment{Run the evolution phase in every iteration}
            
            \If {$i \mod N_\text{update} = 0$}
            \Comment{Update the model every $N_\text{update}$ iterations}
                \State \textsc{UpdateModel($\mathcal{D}$)}
                \Comment{Update the model}
            \EndIf
            
        \EndFor
        \EndProcedure
        
        \Procedure{Bootstrap}{}
        \State $\mathcal{D}_{\text{exploration}} \leftarrow \pi_{\theta}(\alpha | s)_{K}$
        \Comment{Explore the environment $K$ times using random policies and store the data}
        
        \State train($\mathcal{M}$, $\mathcal{D}_{\text{exploration}}$)
        \Comment{Train the VQ-VAE model using the exploration data}
        
        \State $\mathcal{C} \leftarrow \mathcal{M}_{\text{codebook}}$
        \Comment{Set the learned codebook as the initial BD grid}

        \EndProcedure
        
        \Procedure{Evolution Phase}{}
            \State $\boldsymbol{\theta} = $ selection($\mathcal{A}$)
            \Comment{Select randomly solutions from the archive}
            
            \State $\boldsymbol{\theta}' = $ variation($\boldsymbol{\theta}$)
            \Comment{Perform mutation and crossover}
            
            \State $(fitness, d) \longleftarrow $ evaluate($\boldsymbol{\theta}'$)
            \Comment{Evaluate the solution, and get the fitness and the raw data}

            \State $\mathcal{D}^{(i)} \longleftarrow d$
            \Comment{Add the raw data to the training dataset}
            
            \State $\mathbf{bd} \longleftarrow \mathcal{M}_{encoder}(d)$
            \Comment{Extract the behavior descriptor}
            
            \State $c \longleftarrow $ get\_cell\_index($\mathbf{bd}$, $\mathcal{C}$)
            \Comment{Get the index of the cell in the grid the bd is assigned to}
            
            \If {$\mathcal{A}(c) = null$ or $\mathcal{A}(c).fitness < fitness$} 
            \Comment{Check if cooperation mode is on or if the cell is occupied and compete}
                \State $\mathcal{A}(c) \longleftarrow fitness, \boldsymbol{\theta}'$
                \Comment{Add solution to the archive}
            \EndIf
            
        \EndProcedure
        
        \Procedure{UpdateModel}{$\mathcal{D}$}
            \State train($\mathcal{M}$, $\mathcal{D}$) for $\mathcal{E}$ epochs
            \Comment{Update the VQ-VAE model}
            
            \State $\mathcal{C} \leftarrow \mathcal{M}_{\text{codebook}}$
            \Comment{Set the the updated codebook as the BD grid}
            
            \State update\_archive($\mathcal{A}$, $\mathcal{M}_\text{encoder}$)
            \Comment{Update the BDs of the individuals in the archive, and reassign them to the closest cell}
            
        \EndProcedure
    \end{algorithmic}
\end{algorithm*}

A defining feature of the \algo~algorithm is its unsupervised nature, which allows it to autonomously learn behavior descriptors for the discovered solutions. Unlike traditional QD algorithms, where users must manually define BDs~\cite{mouret2015illuminatingsearchspacesmapping, loco_srbd} or rely on prior knowledge to generate BDs using clustering methods~\cite{map-elites-voronoi}, \algo~removes these limitations. Manual or predefined BD selection can restrict the range of problems the algorithm can address or render it impractical in some cases. To overcome this, \algo learns BDs directly from the data.

VQ-VAE is a generative model that combines autoencoders with vector quantization to create discrete latent representations. VQ-VAEs learn a codebook of vectors, mapping continuous latent vectors onto discrete codes. For high-dimensional data such as images, videos, or multivariate time-series, this mapping typically occurs along the channel dimension. However, when applied to one-dimensional data, VQ-VAE performs gradient-based clustering of the latent vectors, where the discrete representations correspond to centroids --- similar to the K-Means algorithm~\cite{lancucki2020robust}. Hence, our algorithm can be used with any data modality, e.g., images, videos, or times-series. However, the latent extraction should be done by a fully connected layer on the flattened features extracted by the previous layers of the encoder.

This clustering property aligns seamlessly with MAP-Elites and the QD framework in general, as it provides both a mechanism for extracting BDs by encoding solution behaviors and a discrete grid (the codebook) to populate with elite solutions. Unlike classical clustering methods, this data-driven approach adapts more effectively to the dynamic nature of the problem, making \algo~significantly more versatile than traditional QD methods.

Alternative unsupervised methods for BD extraction, such as AURORA~\cite{aurora}, face significant challenges when learning from online QD data. The dynamic nature of QD archives means that the behavior space evolves continuously, making it difficult for these methods to adapt effectively in real-time. Moreover, AURORA relies on additional hyperparameters, such as the minimum distance threshold for storing solutions in its unstructured archive. These hyperparameters can be difficult to tune and may require prior domain knowledge, reducing the method's generality and scalability. While AURORA can be effective in certain scenarios, its reliance on unstructured archives and extra hyperparameters makes it less suitable for tasks that benefit from a structured, adaptive behavior space like the one provided by MAP-Elites.

\algo~is split into two phases:
\begin{itemize}
    \item\textbf{Bootstrap phase}: in this phase, we bootstrap the archive and the unsupervised learning process via random policies;
    \item\textbf{Evolution phase}: this is the normal operation of the \algo~algorithm, where every $N_\text{update}$ iterations we update the unsupervised model.
\end{itemize}
\subsection{Bootstrapping}
Initially, the codebook is randomly initialized, something that can slow down the convergence of our algorithm. To overcome this, we first explore the environment to collect some data with random policies. Then, we use the acquired data to bootstrap our codebook by training our model a sufficient amount of epochs, e.g. $\mathcal{E} = 100$. Through this process, we basically warm-start our model and create an initial grid for the evolution phase. In our experience, this step is not mandatory, but it can speed up the convergence of \algo~considerably. Lastly, even though we have not investigated this direction, the user could bootstrap the model with prior information different than exploration data or use a pre-trained model instead.

\subsection{Evolution Phase}
The next key component of \algo~is the evolution phase. During this phase the Quality-Diversity algorithm is executed, performing the usual steps. Foremost, the archive is randomly initialized with random solutions. Then, a random batch of solutions is selected, which are alternated through mutation and crossover. The new variation of the solutions are executed, and evaluated through their behavior descriptor. In order to extract these BDs, we encode the raw data produced by our solutions with the encoder of our VQ-VAE model, acquiring the corresponding continuous latent vectors. For each BD, we find the closest grid cell, and if the cell is empty the solution is added to the archive, otherwise, we compare the fitness of the current solution with the one of the solution that is already in the archive. The solution in the archive is replaced only if the new solution is fitter. This procedure is repeated for a predefined number of iterations. Another important aspect of our framework is that the VQ-VAE model and the MAP-Elites algorithm are completely independent. Hence, the user has the freedom to incorporate different variations of the MAP-Elites algorithm as well as other selection, mutation, and crossover operators.

\textbf{Updating the model:}
The final component of our framework is the model update, refining its latent space and codebook. The update of the model occurs every $N_\text{update}$ iteration and in an online fashion, meaning that the model never resets. Ultimately, the model will converge, and thus, there will be no further changes in the latent space and codebook. To train the model, we exploit the experiences of our agent, by storing the raw data produced by the solutions in the evolution phase. By utilizing these experiences, we can achieve a latent space that encapsulates the underlying dynamics of our agent as well as of the environment. Hence, the behavior descriptors and the grid of our behavior space consist only of feasible behaviors to be produced by the agent, something that would have been impossible to achieve by using a naive clustering algorithm. 

After each update of the VQ-VAE model, we retain a new behavior space, which foreshadows that our current grid and BDs are not aligned. Specifically, this means that the same solution before and after the update of the model is not described by the same BD. To correct this, we need to also update our archive and grid of the behavior space. Consequently, we set the updated codebook as our new grid, and then we reevaluate the solutions we have in the archive, by encoding their raw data and assigning their BDs to the closest cell as in the evolution phase. The update of the archive serves a double purpose; our solutions are accurately represented by the behavior space, and both unfit and less diverse solutions are discarded due to the new refined latent space and codebook.

\subsection{Behavior Space Bounding}
Having discussed the backbone of the \algo~algorithm, we will introduce two key components, which ensure that QD algorithms and unsupervised learning can be coupled effectively, producing high-quality and diverse solutions while performing robustly across tasks. At the core of our approach is the \textbf{manifold hypothesis}, which underpins most representation learning models. This hypothesis suggests that \emph{high-dimensional datasets lie on low-dimensional latent manifolds within the high-dimensional space}. As a result, data points with similar underlying information are clustered closely in the latent space, while dissimilar data points are farther apart.

While this property is advantageous in many contexts, it introduces a potential issue in our framework. Recall that our goal is to learn a behavior space, represented by a latent space, that allows us to find solutions that are both diverse and high-performing. According to the manifold hypothesis, behavior descriptors of diverse solutions will naturally be far apart in the latent space. However, the VQ-VAE model, by design, updates its codebook less effectively than the latent space itself. This disparity can result in an archive filled with high-performing solutions that lack diversity. Additionally, training the model in an online manner also introduces a lot of difficulties, since, until the model converges, newly found diverse solutions are actually out-of-distribution, likely leading to their latent representation not being representative.

To address this challenge, we constrain the latent (and therefore the behavior) space to $[-1,1]$. This bounded space ensures that the latent space remains continuous and expressive, while preventing excessive dispersion that might hinder the codebook’s ability to form a meaningful discretization of the behavior space. By maintaining this bounded structure, we enable the model to produce a diverse set of solutions while preserving high performance.

With this bounded latent space in place, we can further exploit our model to accelerate convergence. Initially, both the VQ-VAE model and its codebook are randomly initialized, which typically requires a significant number of gradient steps to achieve a well-structured discretization. However, given the fixed bounds of the latent space ($[-1,1]$), we can initialize the codebook more effectively. Specifically, we pre-train the codebook by fitting the K-Means algorithm on a sufficient number of samples drawn from the uniform distribution, $\mathcal{U}_{[-1,1]}$.

This initialization provides the codebook with centroids that are already representative of the bounded latent space. As a result, the model requires fewer gradient updates to achieve an optimal discretization of the behavior space. This accelerates both the learning process and the creation of a well-structured archive, ensuring that the algorithm efficiently identifies diverse, high-quality solutions.

\subsection{Cooperation over Competition}
During the evolution phase, new solutions are introduced to the archive by competing with existing ones. However, in the early iterations, the archive often contains less diverse solutions because the representation learning model has not yet fully established a well-defined behavior space. As a result, early individuals in the archive may achieve high fitness scores but lack diversity.

To address this issue, we introduce a cooperation aspect to our algorithm during the initial iterations. This cooperation neglects the competition of individuals, and instead, the grid is filled with the most recent solution assigned to each cell, regardless of fitness. In a way, the individuals are cooperating with offsprings of the population, with the goal of promoting the diversity of the archive. This approach allows the model to focus on constructing a highly diverse grid or refining the behavior space before strict competition begins. Additionally, this strategy enhances the algorithm's exploration capabilities by ensuring frequent updates to the solution set, promoting diversity during the critical early stages of the evolutionary process. \emph{\textbf{We should emphasize that the {\algo} is robust and did not require the cooperation component for our experiments}}, yet we found it critical for other unsupervised QD algorithms we evaluated.
\section{Experiments}
\begin{figure}[!tbh]
    \centering
    \includegraphics[width=0.8\linewidth]{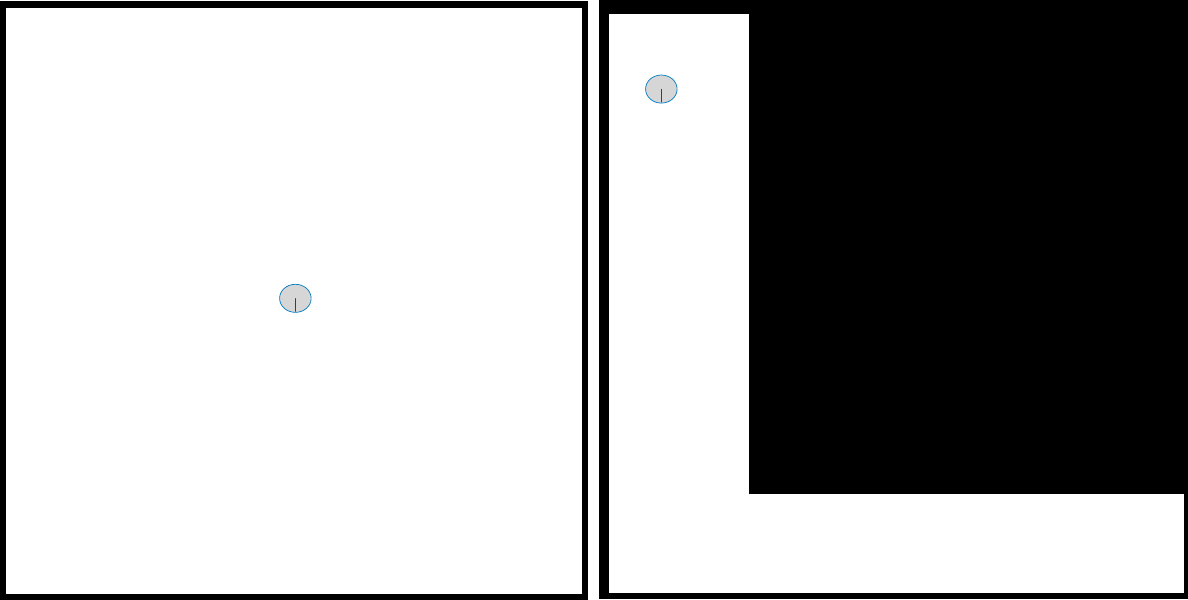}
    \caption{Visualization of the obstacle-free and maze environments for the mobile robot experiments, as well as the starting positions of the robot.}
    \label{fig:mobile_maps}
    \vspace{-1em}
\end{figure}
For the experimental part of our proposed approach, we want to include a representative set of experiments that answer the following questions:
\begin{enumerate}
    \item\textit{How does \algo~compare to the classic MAP-Elites when we have full knowledge of the environment?} 
    \item\textit{How does \algo~performs when we do not possess prior knowledge about the environment or the constraints of the agent?} 
    \item\textit{Is \algo~robust for a range of hyperparameters or requires extensive fine-tuning?} 
    \item\textit{How vital are the behavior space bounding and cooperation components in the convergence of unsupervised QD algorithms?} 
    \item\textit{Can VQ-Elites solve hard exploration tasks?}
    \item\textit{Can VQ-Elites operate with high-dimensional policy structures?}
\end{enumerate}

For our experiments, we selected two state-covering tasks using a mobile robot, two pose-reaching tasks with a 7 degrees of freedom robotic arm, and a hard exploration and discrete task inspired by the MiniGrid Reinforcement Learning environment~\cite{chevalier2018minimalistic}. These tasks were chosen to showcase the versatility and effectiveness of \algo~in solving problems without requiring prior knowledge of the environment or the system. Moreover, these experiments demonstrate the broader potential of \algo~beyond these specific scenarios.

\emph{A key advantage of \algo~is its minimal reliance on hyperparameter tuning. In most of the experiments, the crossover and mutation hyperparameters were kept consistent, as well as the model's update frequency and number of training epochs}. This consistency highlights the robustness and ease of deployment of \algo~across diverse tasks.
\subsection{Mobile Robot Experiments}
For the space-covering experiments, we employ a differential drive mobile robot. The robot is deployed in two environments, one without obstacles and one with a L-shaped environment as depicted in Fig.~\ref{fig:mobile_maps}. During the evolution phase, the algorithm evolves a population that contains the weights of a NN, which acts as the policy. This policy takes as input the state of the robot $\mathbf{s} = \{x, y, \cos(\theta), \sin(\theta)\}$, and produces a $\mathbf{v} = \{v_L, v_R\}$ that includes the velocity that should be applied on each wheel. The goal of this task is for the robot to cover the whole environment while reaching each final position as fast as possible. Hence, the fitness function is:
\begin{equation}
    f = \sum_{i}^N e^{(-\|\boldsymbol{p}_i - \boldsymbol{p}_N\|_2)},
\end{equation}

\noindent where $\boldsymbol{p}_i$ is the position of the robot in the $i$-th step of its trajectory. For every candidate solution, the raw BD is an image of the last position of the robot in the environment with initial dimensions $600 \times 600$, which is then downscaled to $64 \times 64$ and fed into the encoder network of VQ-VAE to extract our low-dimensional BD. Additionally, these images are stored for the learning phase of our pipeline. We should note, that the reward of the environment is only responsible for the fitness of the solution. The space-covering aspect of the task is fulfilled in the learning phase, and thus, the model should construct an optimal desensitization of our behavior space to achieve it.

During the learning phase, we noticed that the model became biased due to the constructed dataset being unbalanced. This phenomenon arises as a combination of two things, the ratio of important information in the images is low and at the beginning of our algorithm, the solutions are not diverse enough. To overcome that, we filter our dataset by removing instances that are very similar to each other, by checking the amount of overlap the robot has on the pixel level. 

Lastly, the initial position of the robot is fixed, and the algorithm is executed for 3000 iterations, while each solution performs 400 steps in the environment.  

\begin{figure}[!tbh]
    \centering
    \includegraphics[scale=0.30]{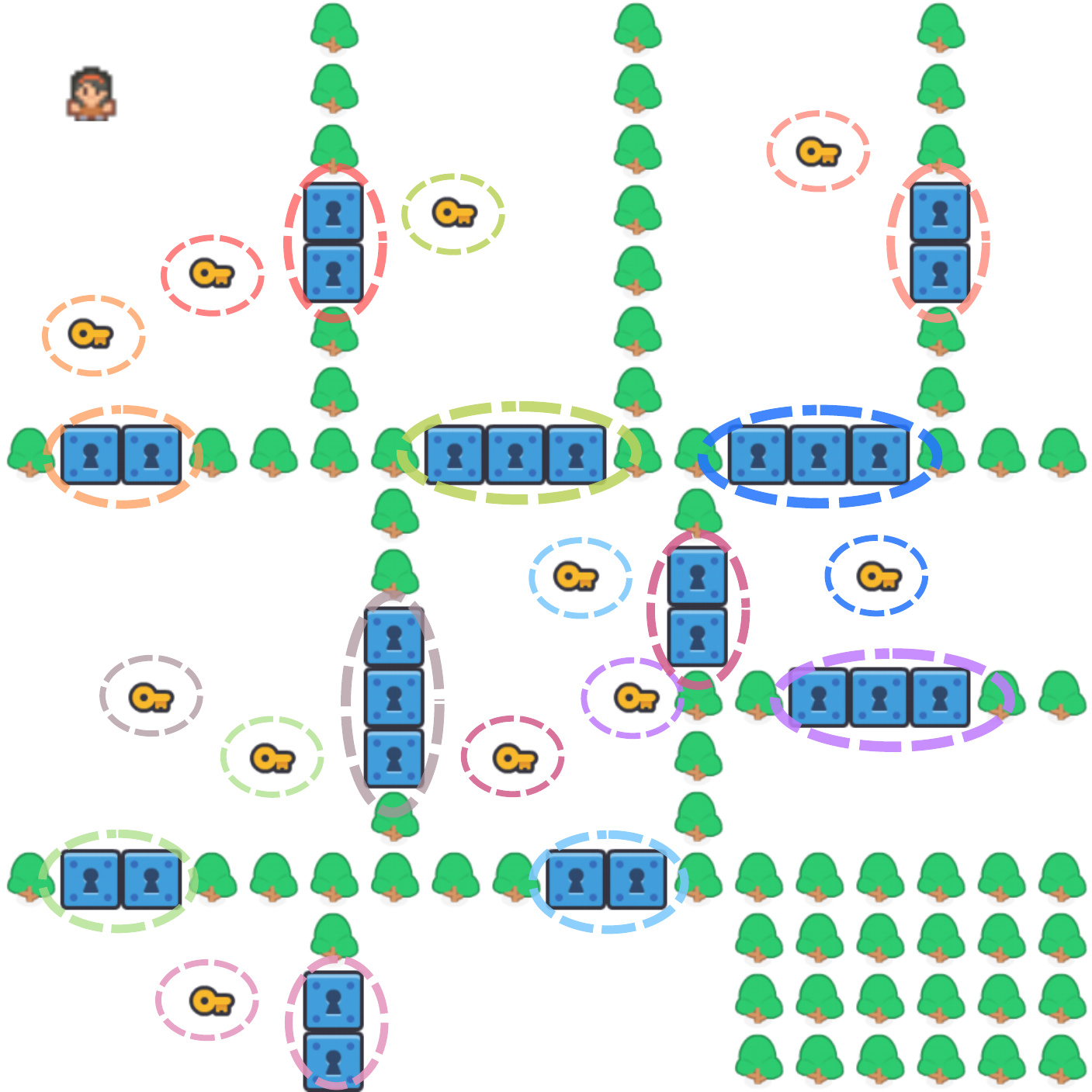}
    \caption{Overview of the MiniGrid task's map. The color-coded circles around the keys and the doors, represent key-door pairings.}
    \label{fig:minigrid_map}
\end{figure}
\subsection{Robotic Arm Experiments}

Regarding the pose-reaching tasks, the general notion of the pipeline follows the mobile robot experiment, with some variations. The goal of the task is for the end-effector of the robot to reach a predefined location, with as diverse joint configurations as possible. Here, the input of our policy as well as the data we use to extract the BD and train the VQ-VAE model are the joints of the robot (raw BD). The output of the policy is the joint velocities. The fitness formula for the robotic arm is:

\begin{equation}
    f = e^{(-\|E_{g} - E_{N}\|_2)}
\end{equation}
\noindent where $E_{g}$ is the 3D coordinates of the goal, and $E_{N}$ is the 3D coordinates of our end-effector at the last step of the rollout.

In order to demonstrate the flexibility of our algorithm, we first showcase this experiment with the default limits of the joint space, and after, we introduce some artificial constraints to the robot, which simulate a problematic agent or a confined operation space. The goal of this variation is to exhibit the high potential of unsupervised learning in combination with QD algorithms in solving problems in an agnostic manner, where the constraints of the system cannot be identified beforehand. For both of these tasks, we use 3000 iterations, and each policy performs 300 environment steps.

\subsection{MiniGrid Exploration}

The MiniGrid experiment (Fig.~\ref{fig:minigrid_map}) consists of a grid map, and an agent moving (with discrete actions) inside it. We use this experiment to investigate the performance of our approach in a domain other than robotics. The ultimate goal of the task is to discover individuals through the evolution process to traverse all possible tiles of the map. The agent has a field of view (FOV) of 3 tiles and can perform 5 actions, $\mathbf{A} = $\{\textit{up, right, down, left, stay still}\}, which are selected by the policy. For the policy's input, we use snapshots of the current agent's position while respecting its FOV, along with the list of keys that have been gathered using one-hot encoding. Hence, these snapshots contain information in pixels inside the FOV of the agent, while the rest are masked out. Additionally, the snapshot of the last step in the environment also serves as our raw BD. The size of the snapshots is $18\times18$.

At every episode, the agent needs to solve two subtasks multiple times in order to get access to the rest of the map. Firstly, it needs to identify any keys present and reach their tiles to grab them. Secondly, it needs to find the doors that correspond to those keys to unlock them. By solving these subtasks, the agent is rewarded with 10 points each time it finds a key and with 20 points each time it opens a door. The accumulated points of each rollout act as our fitness.

The two key characteristics that make this experiment interesting are the discrete nature of the environment and its hard-to-explore nature. Compared to the previous experiments, where they are governed by smooth dynamics and transitions, the discrete nature of the current experiment showcases that our algorithm is agnostic to the dynamics of the task at hand. Additionally, this task is a hard-to-explore problem, both in the state space and solution space. To make the state space exploration difficult, we designed a map where the agent needs to solve multiple rooms to traverse the whole map (Fig. \ref{fig:minigrid_map}). On the other hand, the policy we deploy is a CNN network with a high number of parameters ($\approx 3$K), which makes the evolution more hectic and the discovery of optimal solutions harder. We should note that on Fig~\ref{fig:minigrid_map}, the graphics are only for visual presentation. In the map used for our experiment, the key, doors, obstacles, and agent are represented by uniquely colored tiles for each item. Overall, for this experiment, we perform 10,000 iterations, and 200 environment steps per evaluation.

\subsection{Evaluation Metrics}

To evaluate the performance of our algorithm, we employ most of the metrics used in AURORA \cite{aurora}, but we also introduce two new metrics. Specifically, we evaluate our approach in terms of coverage, (Projected) QD score, and our newly proposed Effective Diversity Ratio and Coverage Diversity Score. We also introduce a visualization method for the unsupervised setting, where the algorithms' archives are projected onto a grid defined by the ground-truth behavior descriptor. We refer to this visualization as the \emph{Elite Grid Projection}.

\subsubsection{Coverage}

This metric measures the percentage of the hand-coded behavior space that is covered. The BD space is divided into a grid of cells, and the coverage is calculated as the proportion of cells containing at least one individual. In the mobile robot experiments, since we operate in a 2D space, we simply discretize the robot's operating space in cell of dimension $N \times N$. Similarly, in the MiniGrid case, since the state is discrete, we can easily measure the ground truth coverage. On the other hand, this is not possible in the pose-reaching task, hence, we gather 1M poses using inverse kinematics (IK) and perform a K-Means clustering on these data. The resulting centroids correspond to our hard-coded BD.

\subsubsection{Projected Quality-Diversity Score}
The QD score is used to assess the performance of an algorithm in terms of both the diversity and the quality of solutions in the BD space. It is calculated as the sum of the fitness values of all the individuals across the grid. By combining diversity (coverage of the grid) and quality (fitness of solutions), the QD score provides a single, unified measure to evaluate the effectiveness of a given algorithm. Higher QD scores indicate better exploration and exploitation of the BD space. However, for the QD score to be meaningful, it requires all the algorithms to have the same behavior space grid, which is impossible in an unsupervised setting. Hence, we use the Projected QD (PQD) score, which is a proxy of the QD score. The PQD score is calculated by discretizing the ground truth behavior space into cells, for each cell, we keep the elite, and a new \emph{projected archive} is generated. Then, by calculating the classic QD score on this newly generated projected archive, we get the PQD score.

\subsubsection{Effective Diversity Ratio}
We introduce the Effective Diversity Ratio (EDR), a novel metric designed to enable fairer comparisons between unsupervised QD algorithms. EDR quantifies the proportion of genuinely diverse solutions discovered by an algorithm, by measuring the ratio of unique behavior outcomes to the total number of valid individuals in the archive. To compute EDR, we discretize the behavior space into $N$ bins, where $N$ represents the number of desired distinct behaviors. Each individual in the archive is evaluated by executing it in the environment (or task), and the resulting solution, i.e., the observed behavior outcome, is assigned to the nearest bin in the discretized space. The EDR is then defined as the number of occupied bins in the ground truth grid divided by the total number of valid individuals in the original archive:
\begin{equation*}
    \text{EDR} = \frac{\text{Number of Occupied Bins}}{\text{Valid Archive Size}}
\end{equation*}
\noindent This metric reflects how effectively the algorithm explores the behavior outcome space, discounting redundancy among individuals whose executions lead to similar behaviors. A higher EDR indicates more unique behaviors and less redundancy in the discovered solutions.

\subsubsection{Coverage Diversity Score}
We also introduce the Coverage Diversity Score (CDS), a metric designed to assess behavioral diversity in the context of archive coverage in the unsupervised setting. It is defined as the product of Coverage and EDR:
\begin{equation*}
    \text{CDS} = \text{Coverage}\; \times\; \text{EDR}
\end{equation*}
\noindent While EDR represents the inverse redundancy in the archive, it does not always correspond directly to meaningful diversity. For example, at the very beginning of evolution, an archive containing a single individual will yield an EDR of $100\%$, despite offering no meaningful diversity. Conversely, as the archive grows, the EDR may become artificially low even if the behavior spread remains satisfactory.
By incorporating coverage, the CDS score mitigates these issues, providing a more balanced view of diversity relative to the size of the explored behavior space. It offers a more robust indicator of how effectively an algorithm populates the behavior space with both broad and distinct solutions.

\begin{table}
 \caption{List of all the hyperparameters used for each type of environment}
 \label{tab:hyperparameters}
 \centering
\begin{tabular}{ l c c c }
 \toprule
 \multicolumn{4}{c}{Experiment Hyperparameters} \\
 \midrule
 Experiment & Mobile robot    & Robotic arm  & MiniGrid \\
 \midrule
 Iterations   & $3 \times 10^3$    & $3 \times 10^3$ & $10^4$ \\
 Population size & $128$ & $128$ & $128$ \\
 $|\mathcal{C}|$ \scriptsize(target archive size) &   $2000$  & $1500$ & $400$ \\
 Raw BD size &   $[1, 64, 64]$ & $6$ & $[3, 18, 18]$  \\
 BD/Latent size &   $2$  & $5$ & $5$ \\
 $\mathcal{E}$ &   $10$  & $10$ & $10$ \\
 $lr$ &   $7\times10^{-4}$  & $7\times10^{-4}$  & $7\times10^{-4}$ \\
 Batch size &   $64$  & $64$  & $64$ \\
 $N_\text{update}$ &   $5$  & $5$  & $10$ \\
 $N_\text{cooperation}$ & --  & -- & --  \\
 \midrule
 \multicolumn{4}{c}{Evaluation Hyperparameter} \\
 \midrule
 Projection Grid Cells & $30 \times 30$ & 400 & $18\times18$ \\
 \bottomrule
\end{tabular}
\end{table}
\section{Results}

In this section, we showcase the results of our algorithm in the experimental evaluation, starting with the mobile robot experiments, then continuing to robotic arm experiments and the MiniGrid exploration task, and finally presenting ablation studies we performed on our algorithm. We report all the hyperparameters used in each setting at Table~\ref{tab:hyperparameters}. For the AURORA-based methods, we used the same environment-specific hyperparameters (target archive size, population, etc.) as for {\algo}, and followed the values of the original paper for the rest hyperparameters~\cite{aurora}\footnote{We also refer to the supplementary material for the complete list of hyperparameters.}.

\subsection{Mobile Robot Results}
For the mobile robot experiments, we compare {\algo} algorithm with the classic MAP-Elites algorithm, the AURORA approach, as well as the AURORA after integrating our two proposed components, behavior space bounding and cooperation, which from now on we will refer to as {\aur}. We should mention that a variation of this experiment is also present in the AURORA paper~\cite{aurora}, hence it is ideal for comparison purposes, however, we are unable to exactly replicate the authors' results\footnote{We used the exact hyperparameters presented in the original paper and communicated with the authors.}, as demonstrated in the manuscript. 

\begin{figure}
    \centering
    \includegraphics[width=\linewidth]{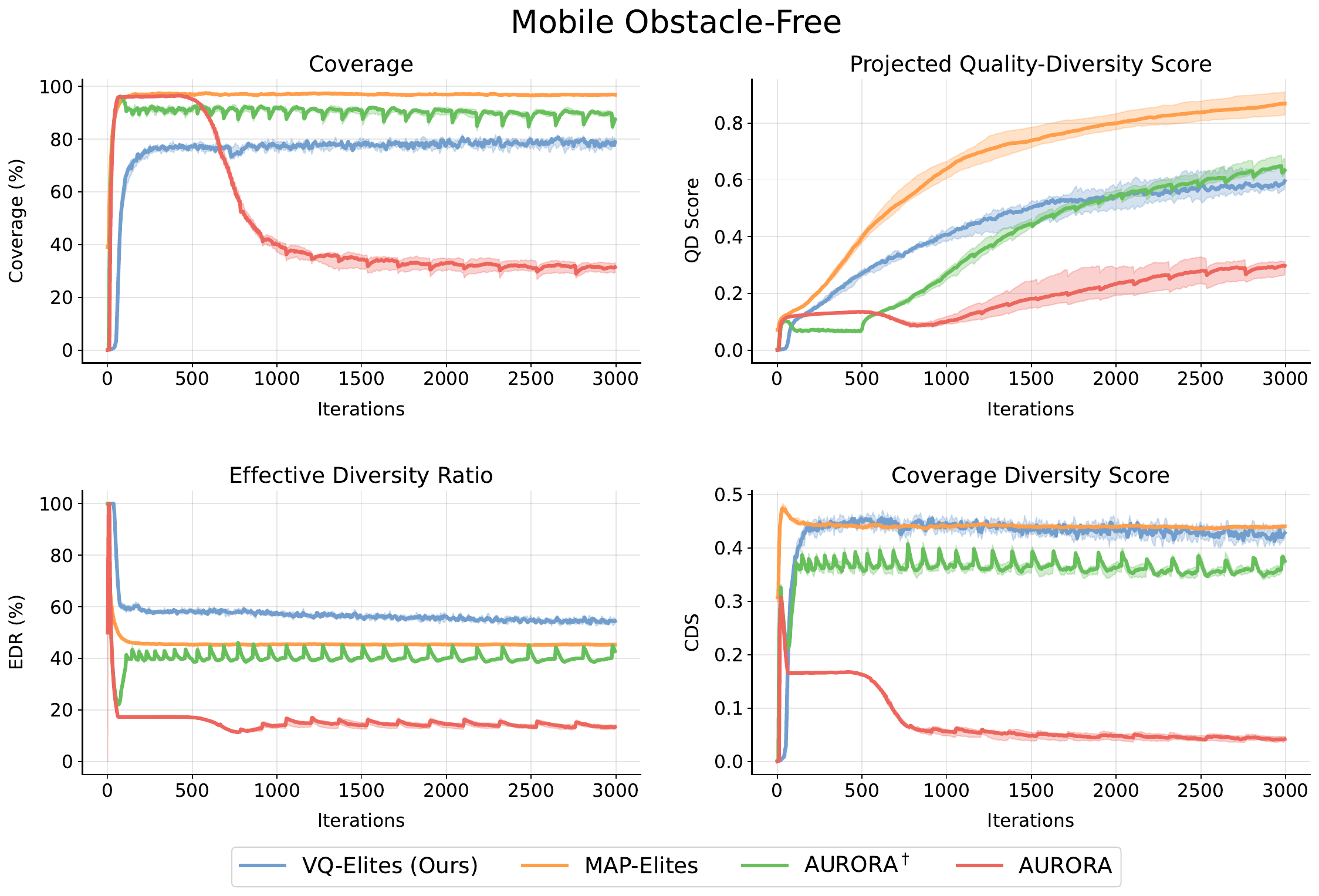}
    \caption{Results for the mobile robot operating in the obstacle-free environment of four different algorithms, {\algo}, AURORA, {\aur}, and MAP-Elites with hard-coded behavior descriptors. Each graph line represents the median and the 25th and 75th quantile over ten runs for the corresponding algorithm.}
    \label{fig:mobile_full_result}
\end{figure}

In the case where the robot operates in an obstacle-free environment, we can observe that all the algorithms, except the vanilla AURORA, perform almost perfectly in terms of coverage, assigning solutions to nearly every grid cell (Fig.~\ref{fig:mobile_full_result}). In terms of the PQD score, we see that the classic MAP-Elites algorithm produces more high-quality and diverse solutions compared to the other approaches. Both of these observations are expected, since we can easily define a behavior grid manually, and still every cell will represent a feasible behavior. Nonetheless, it is evident that algorithms that assume no prior knowledge about the environment, such as ours and other unsupervised QD algorithms infused with our cooperation and behavior space bounding components, can achieve the performance of MAP-Elites without full knowledge of the task (answers to Question~1). Additionally, considering the EDR and CDS metrics, {\algo} is producing new behaviors more effectively compared to {\aur} and it is on par with MAP-Elites. This indicates that despite the coverage of {\algo} being lower compared to the other two, the behaviors produced do a better job of promoting diversity while maintaining a high-quality archive.

\begin{figure}
    \centering
    \includegraphics[width=\linewidth]{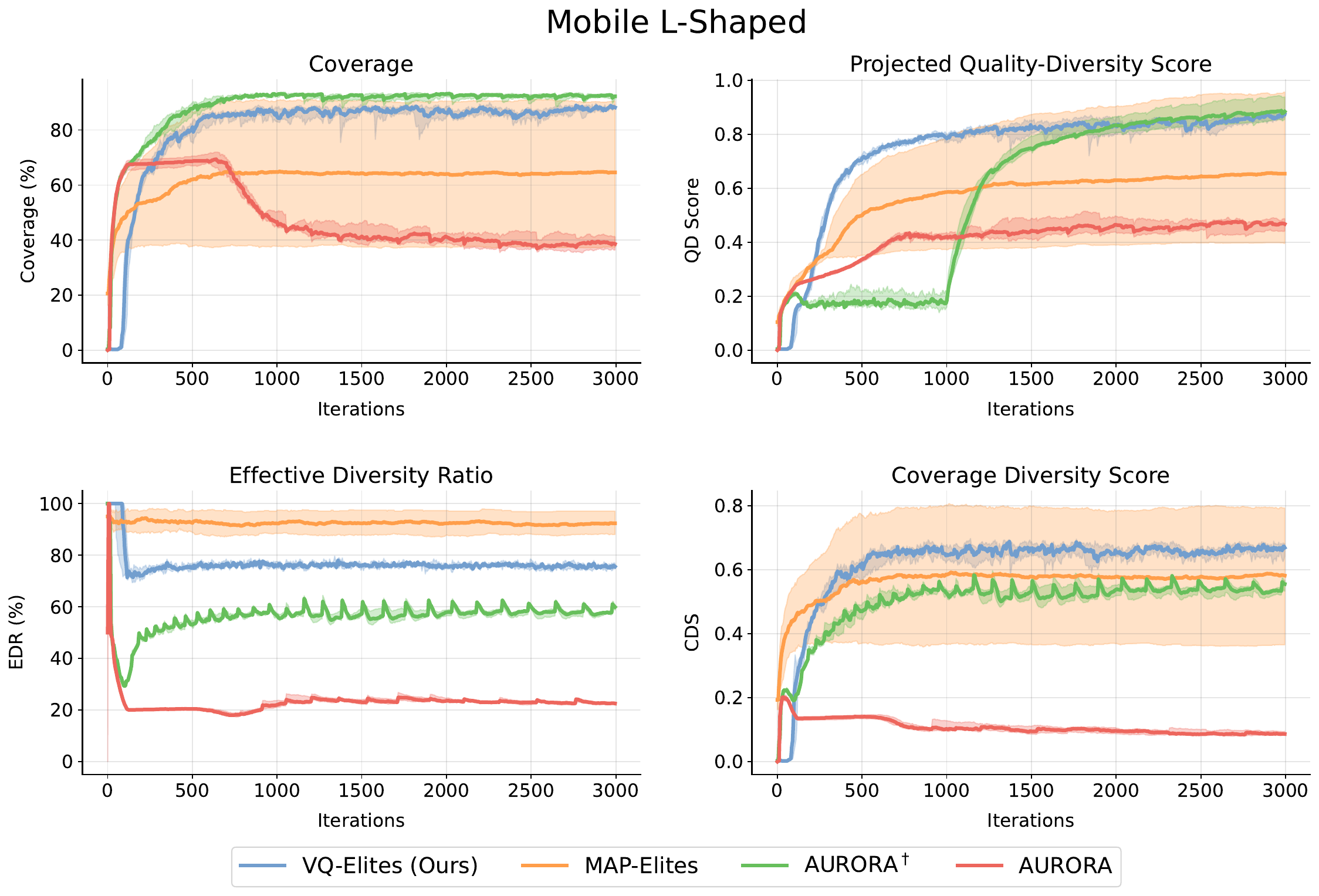}
    \caption{Results for the mobile robot operating in the L-shaped environment of four different algorithms, {\algo}, AURORA, {\aur}, and MAP-Elites with hard-coded behavior descriptors. Each graph line represents the median and the 25th and 75th quantiles over ten runs for the corresponding algorithm.}
    \label{fig:mobile_L_results}
\end{figure}

Transitioning to the L-shaped environment, our approach and {\aur} outperform the others by a large margin in terms of coverage, while {\algo} exhibits the best PQD score of all (Fig.~\ref{fig:mobile_L_results}). This is one type of task that makes the existence of unsupervised QD algorithms essential, since, if we do not possess full knowledge about the task at hand, and particularly in this scenario, we do not know the environment, we cannot construct a grid such that every cell corresponds to a possible behavior (answers to Question~2). On the other hand, a data-driven unsupervised approach is more than capable of constructing such a grid. Although MAP-Elites has low coverage, it outperforms the other algorithms in the effective diversity metric. This is expected since the hand-crafted behavior grid is well-discretized, enabling the algorithm to provide a small but diverse archive. However, this is not the case in the CDS metric, where {\algo} has achieved a higher score since the coverage is taken into account. Additionally, between {\algo} and {\aur}, our algorithm performs better in the EDR and CDS metrics, since it does not rely on an unstructured archive, which populates the archive based on heuristics that may result in many duplicate behaviors.

\begin{figure*}[h!]
    \begin{subfigure}{0.5\linewidth}
  \includegraphics[scale=0.4]{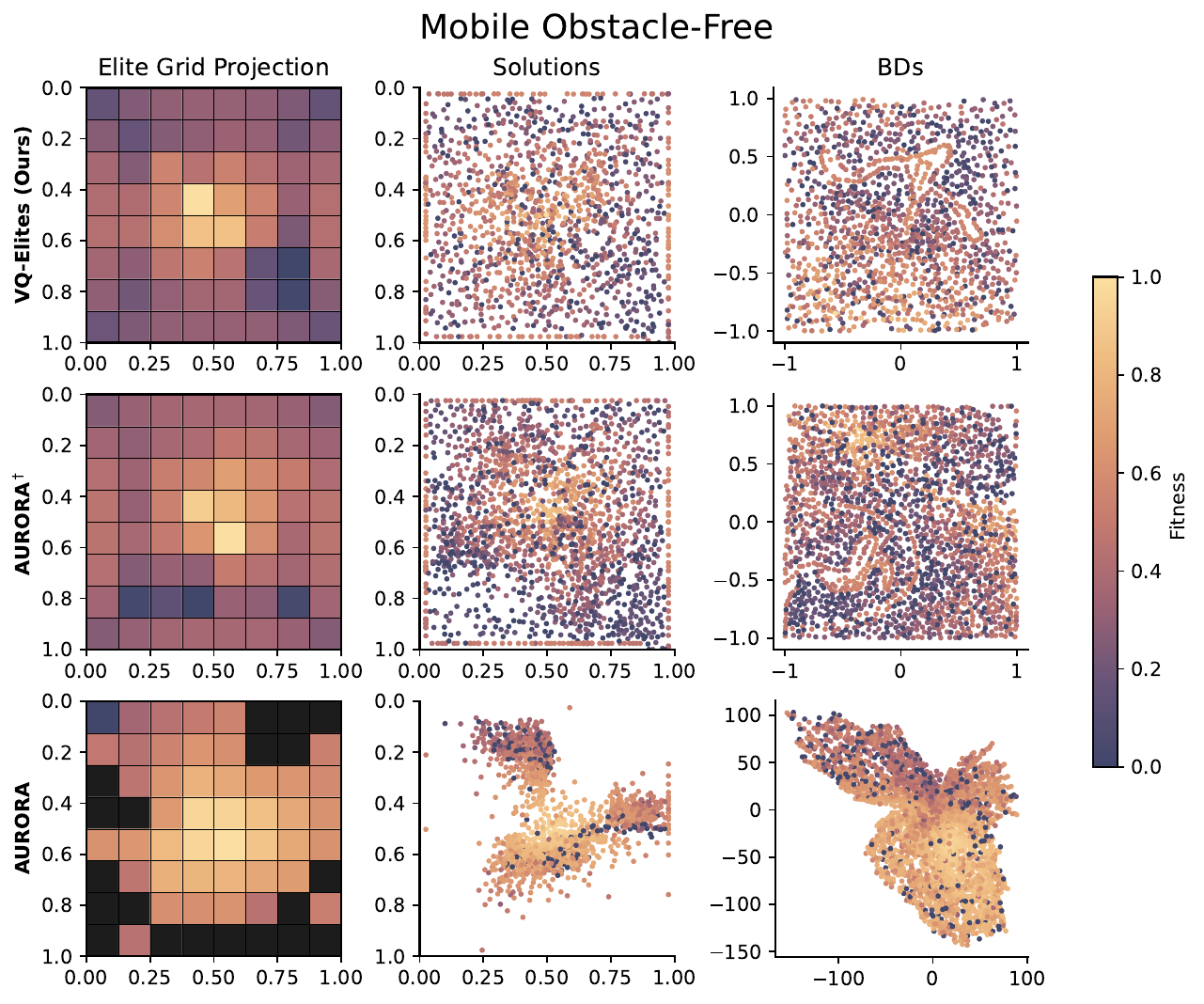}
  \end{subfigure}
    \begin{subfigure}{0.5\linewidth}
  \includegraphics[scale=0.4]{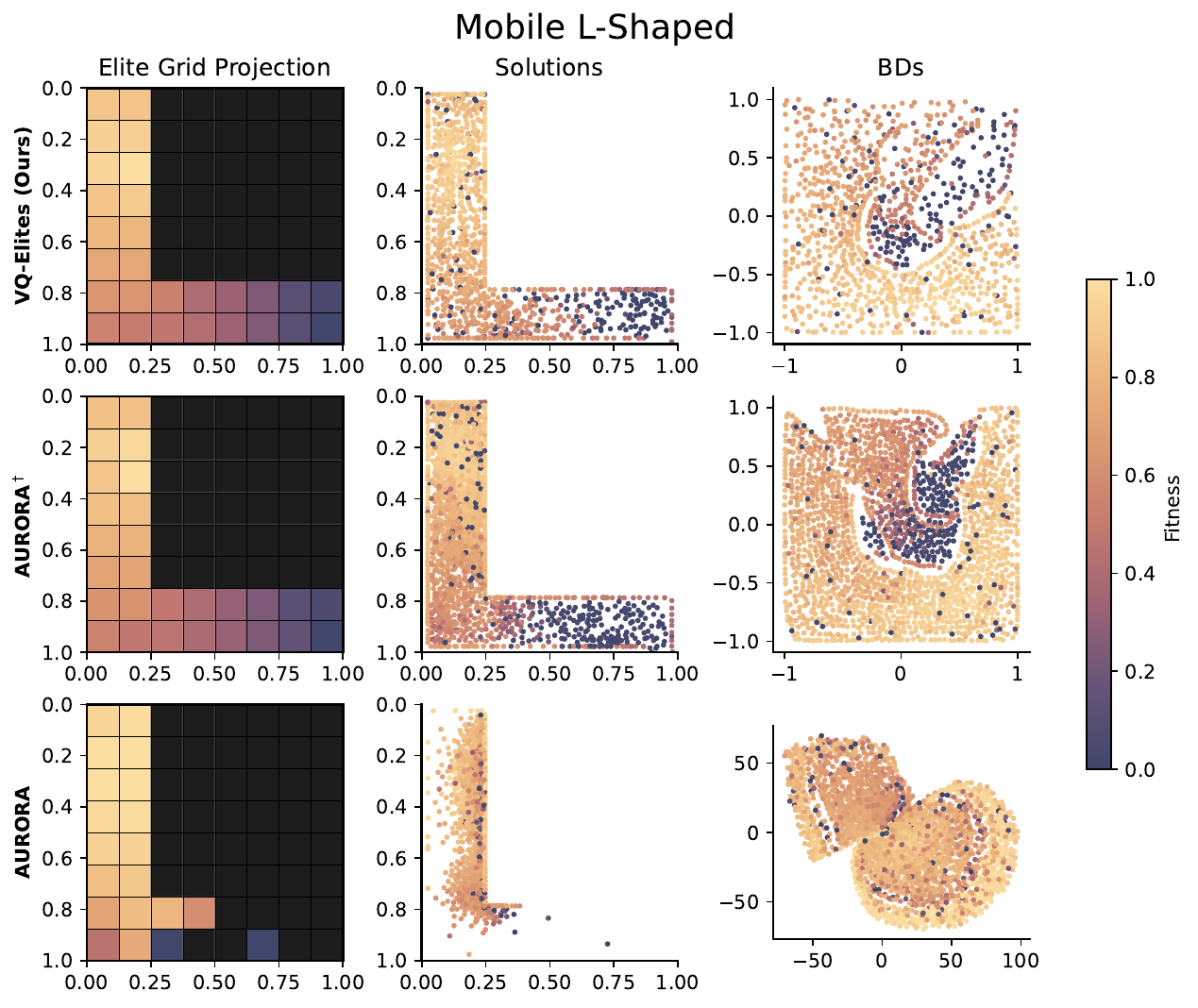}
  \end{subfigure}
    \caption{Visualizations that demonstrate the mobile experiment's outcomes of the three algorithms in the two different environments. \textit{Left}: Obstacle-free experiment outcome. \textit{Right}: L-Shaped experiment outcome. In both plots, the first column indicates the elite grid projection, where we discretize the solution space in an $8\times8$ grid and keep the fitness of the elite individual. The second depicts where every learned behavior ends up in the environment. The third column is a visualization of the behavior descriptors that describe these behaviors.}
    \label{fig:mobile_all_aurora}
\end{figure*}

Lastly, since the experiment is two-dimensional, we also provide plots, which are hard or infeasible to do or comprehend in high dimensions. These plots depict the elite grid projection, final archives of the algorithms, and the behavior descriptors for each algorithm per scenario in Fig.~\ref{fig:mobile_all_aurora}. In the first column of each plot, we illustrate the elite grid projection by discretizing the robot's operating space into an $8\times8$ grid, and the color of each cell represents the normalized fitness of the best solution assigned to each cell. The second column showcases the actual solutions {\algo} produced, while the last one depicts the BDs of the solutions that emerged. Overall, we observe that {\algo} and {\aur} are able to effectively explore the behavior space, produce meaningful latent representations, and in the case of {\algo}, it automatically performs clustering of the behaviors, constructing a meaningful behavior space grid.

\begin{figure*}[!tbh]
      \centering
      \centerline{\includegraphics[width=\linewidth]{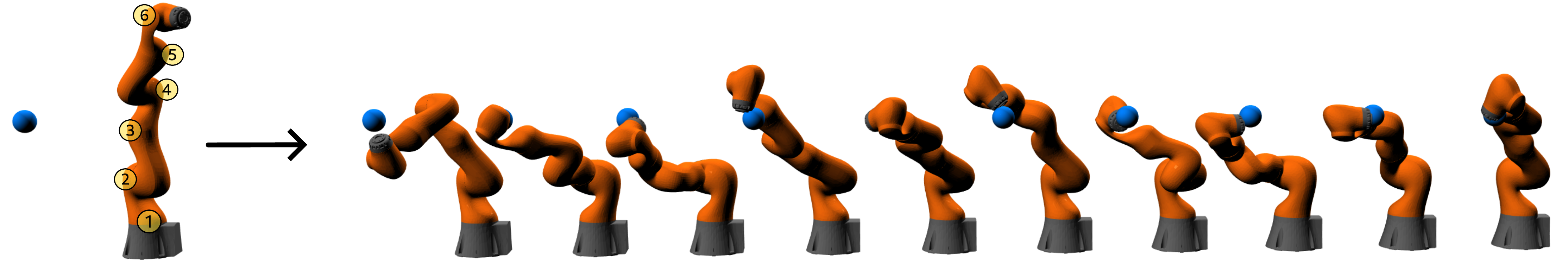}} 
    \caption{Illustration of a subset of diverse poses (individuals) achieved from the {\algo} algorithm. The blue sphere indicates the end-effector target. \textit{Left}: Robot's starting pose with indicators about each joint's position. \textit{Right}: Learned behaviors.}
    \label{fig:poses}
\end{figure*}
\subsection{Robotic Arm Results}
For the robotic arm experiments, we require the robot to achieve diverse poses, filling the whole operation space. In Fig.~\ref{fig:poses}, we provide a visualization of some diverse poses our algorithm achieves. For this task, we investigate two different aspects, firstly, how the structured grid archive {\algo} uses compares to the unstructured archive of {\aur}. Secondly, if the robotic arm system has some constraints, e.g. due to motor malfunction, and thus some joints are not fully operational, can our algorithm still find optimal solutions, and how does it compare to the classic MAP-Elites algorithm?

\begin{figure}[h]
    \centering
    \includegraphics[width=\linewidth]{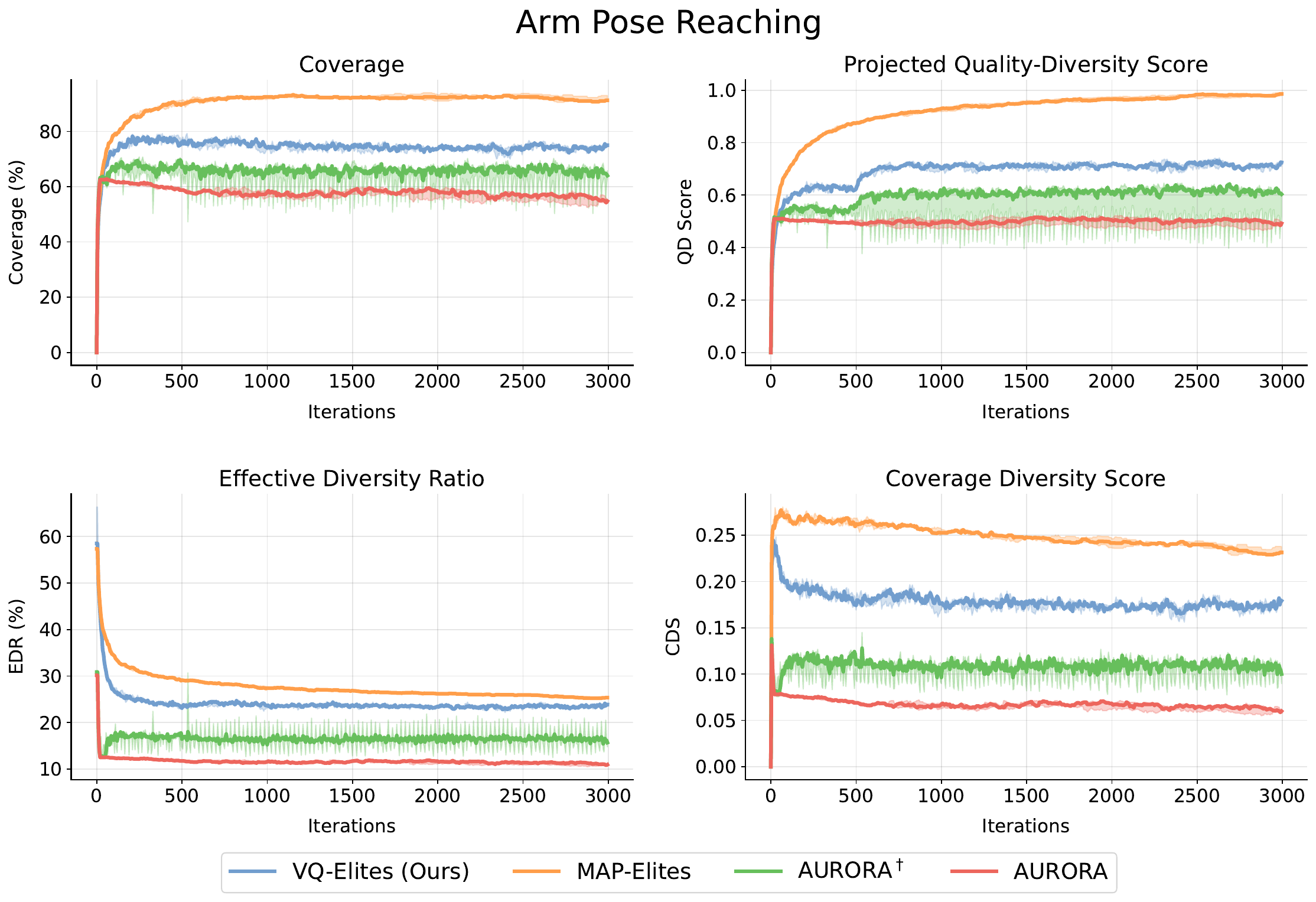}
    \caption{Results for the robotic arm pose reaching task for {\algo}, MAP-Elites, AURORA, and {\aur}. Each graph line represents the median and the 25th and 75th quantiles over ten runs for the corresponding algorithm.}
    \label{fig:robotic_arm_simple}
\end{figure}

For the first scenario, it is evident that MAP-Elites is the best of the batch in both coverage and PQD score. {\algo} and {\aur} follow, while {\algo} is slightly better (Fig.~\ref{fig:robotic_arm_simple}). This fact indicates that the way an unstructured archive is updated is highly likely to discard high-quality and diverse individuals in order to satisfy the minimum distance between individuals rule~(answers to Question~1). Out of the four algorithms, vanilla AURORA performs the worst.

\begin{figure}[!tbh]
    \centering
    \includegraphics[width=\linewidth]{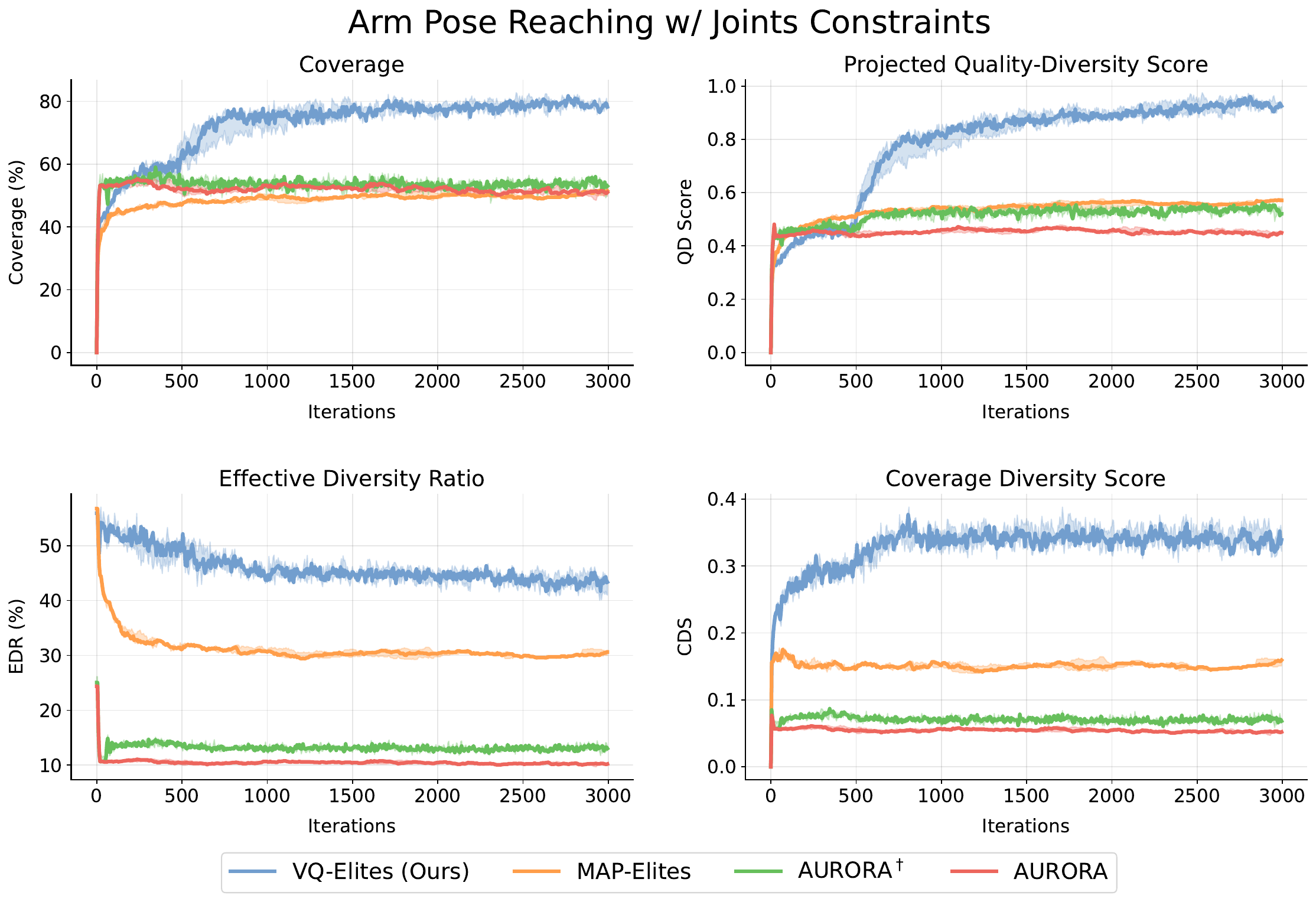}
    \caption{Results for the robotic arm pose reaching task with system constraints (i.e. limitation in some joints) for {\algo}, MAP-Elites with hard-coded BDs, AURORA, and {\aur}. Each graph line represents the median and the 25th and 75th quantiles over ten runs for the corresponding algorithm.}
    \label{fig:robotic_arm_constrained}
    \vspace{-1em}
\end{figure}

The second case resembles the mobile robot in the L-shaped environment. However, in this task, the constraint is on the robot itself, by reducing some joint limits. Therefore, we change the limits of joints 1 and 2 (see Fig. \ref{fig:poses}) from $[-2.967, 2.967]$ to $[-0.5, 0.5]$. Despite that, our algorithm has a completely agnostic nature due to the unsupervised learning model, and hence, it outperforms all the other approaches by a large margin in both coverage and PQD score (Fig.~\ref{fig:robotic_arm_constrained}). We should note that this is highly predictable because in order for the MAP-Elites to achieve a competitive performance, it requires knowledge of the constraints to construct a proper behavior descriptor grid. Lastly, both {\aur} and vanilla AURORA perform similarly to the MAP-Elites algorithm despite their unsupervised nature, while the PQD score of the vanilla variation is the worst of the three due to the way the archive is constructed, as noted previously  (answers to Question~2). Focusing on the effective diversity and coverage diversity score, our algorithm is still outperforming the others. For instance, the $50\%$ of {\algo}'s archive is filled with unique behaviors in the former metric while achieving the best score over all algorithms in the latter. Surprisingly, the {\aur} and vanilla AURORA fail to provide a competent amount of diverse behaviors.

\subsection{MiniGrid Results}
\begin{figure}[!tbh]
    \centering
    \includegraphics[width=\linewidth]{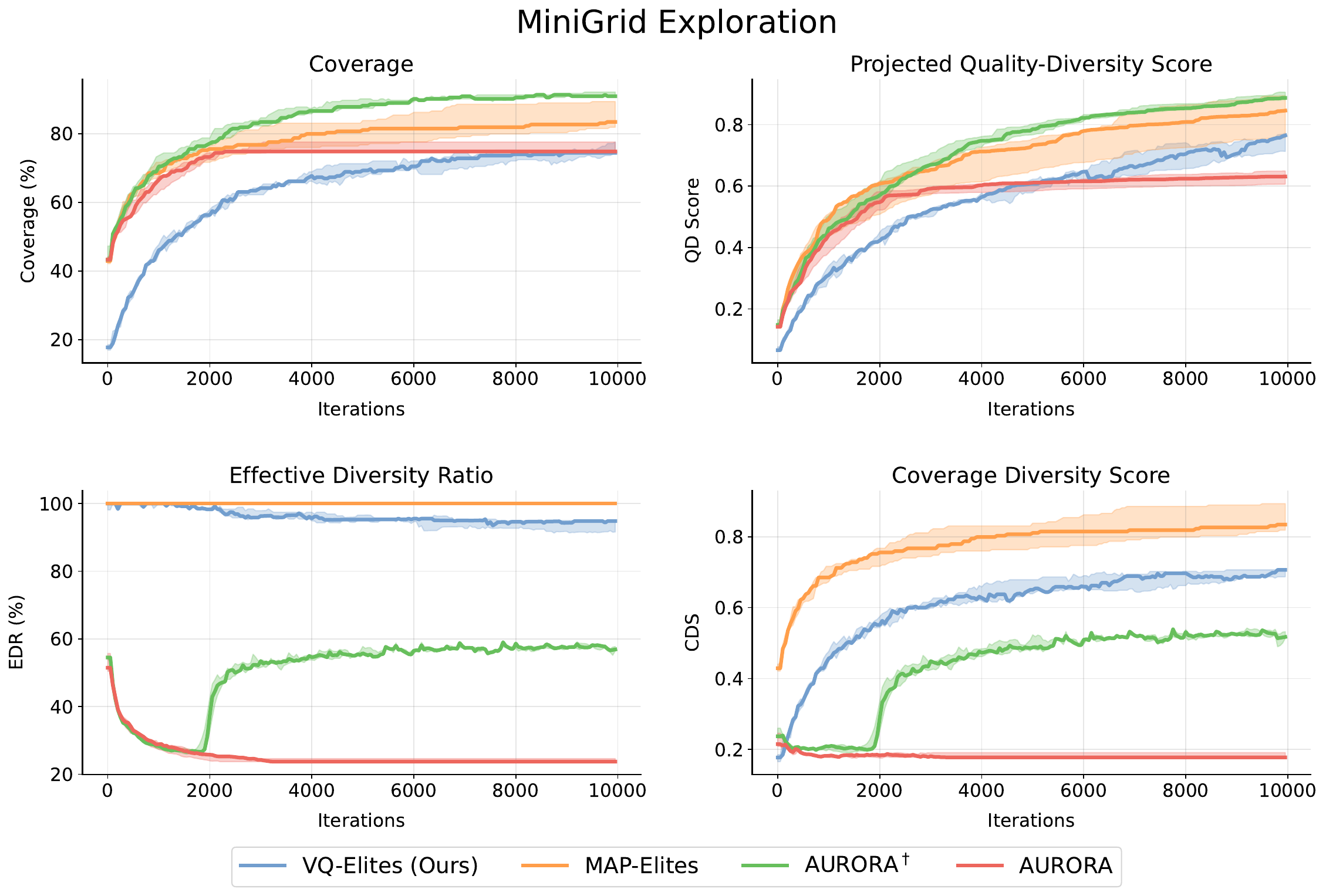}
    \caption{Results for the MiniGrid exploration task for {\algo}, MAP-Elites with hard-coded BDs, and {\aur}. Each graph line represents the median and the 25th and 75th quantiles over five runs for the corresponding algorithm.}
    \label{fig:minigrid_coverage}
\end{figure}

\begin{figure}[!htb]
    \centering
    \includegraphics[width=\linewidth]{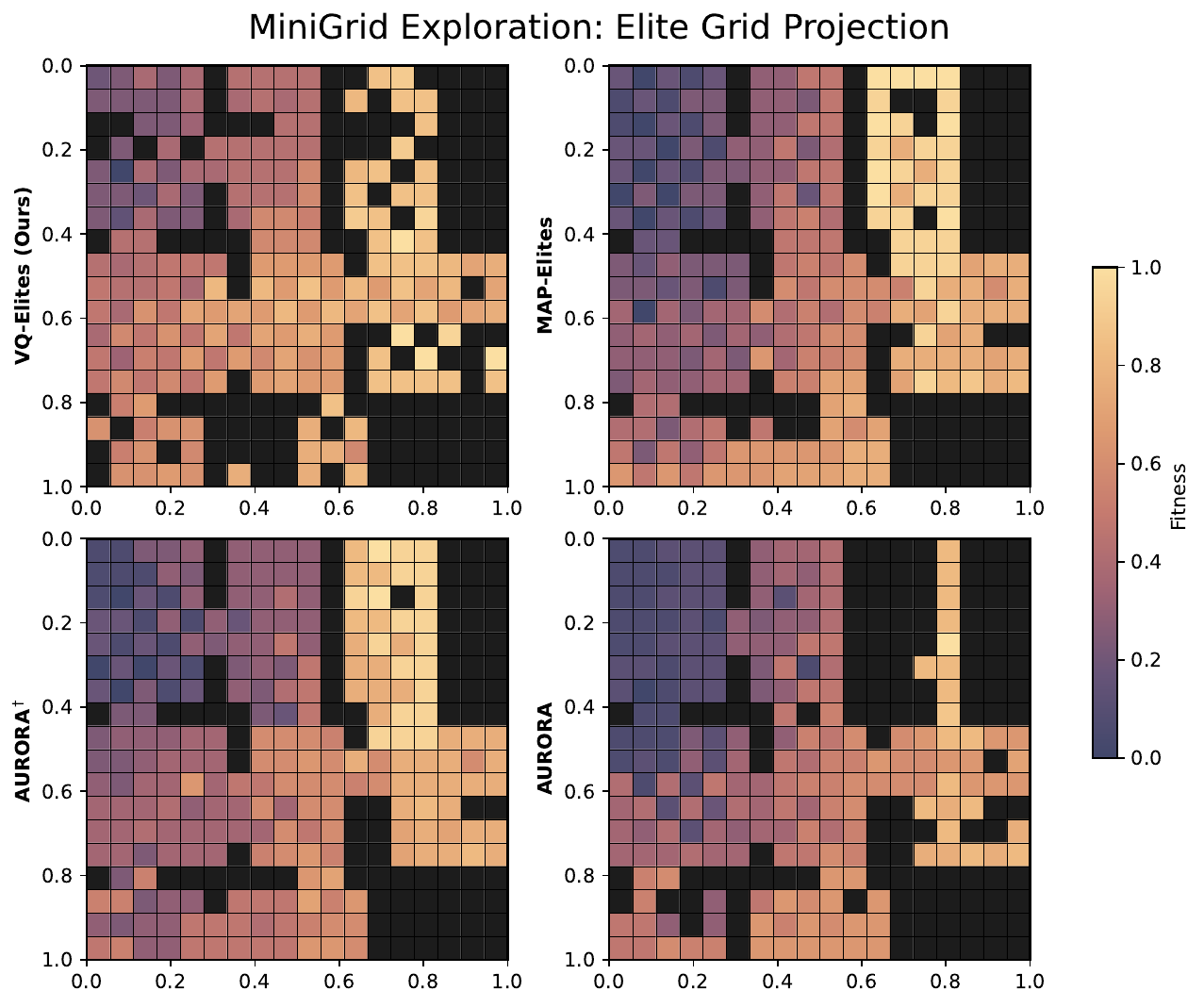}
    \caption{Elite Grid Projection of the MiniGrid exploration task for {\algo}, MAP-Elites with hard-coded BDs, and {\aur}}
    \label{fig:minigrid_grid_fit}
\end{figure}

In the MiniGrid experiment, we evaluate the performance of the algorithms in a discrete and challenging exploration setting. The results reveal that convergence of {\algo} in terms of coverage is significantly slower than in previous tasks (Fig.~\ref{fig:minigrid_coverage},~\ref{fig:minigrid_grid_fit}), but it outperforms by quite some margin the other unsupervised algorithms on the effective diversity and coverage diversity metrics~(answers to Questions 5 and 6). In contrast, both variants of the AURORA algorithm converge faster but fail to produce a truly diverse archive.

This difference in performance is primarily due to the use of the unstructured archive in AURORA. Such archives can be advantageous in difficult exploration scenarios when coverage alone is the primary objective. Their lack of structure allows a large number of individuals to persist in the population, effectively "brute-forcing" the evolutionary search. This increases the pool of candidates available for crossover and mutation, which in turn improves the likelihood of producing individuals capable of better traversing the environment. This effect is particularly pronounced in the MiniGrid setting, where the high-dimensional policy space requires significantly more exploration than in the previous tasks. As a result, unstructured methods can gain an advantage in raw coverage at the cost of behavioral diversity.

\subsection{Ablation Study}

\begin{figure}[h]
    \centering
    \includegraphics[width=0.9\linewidth]{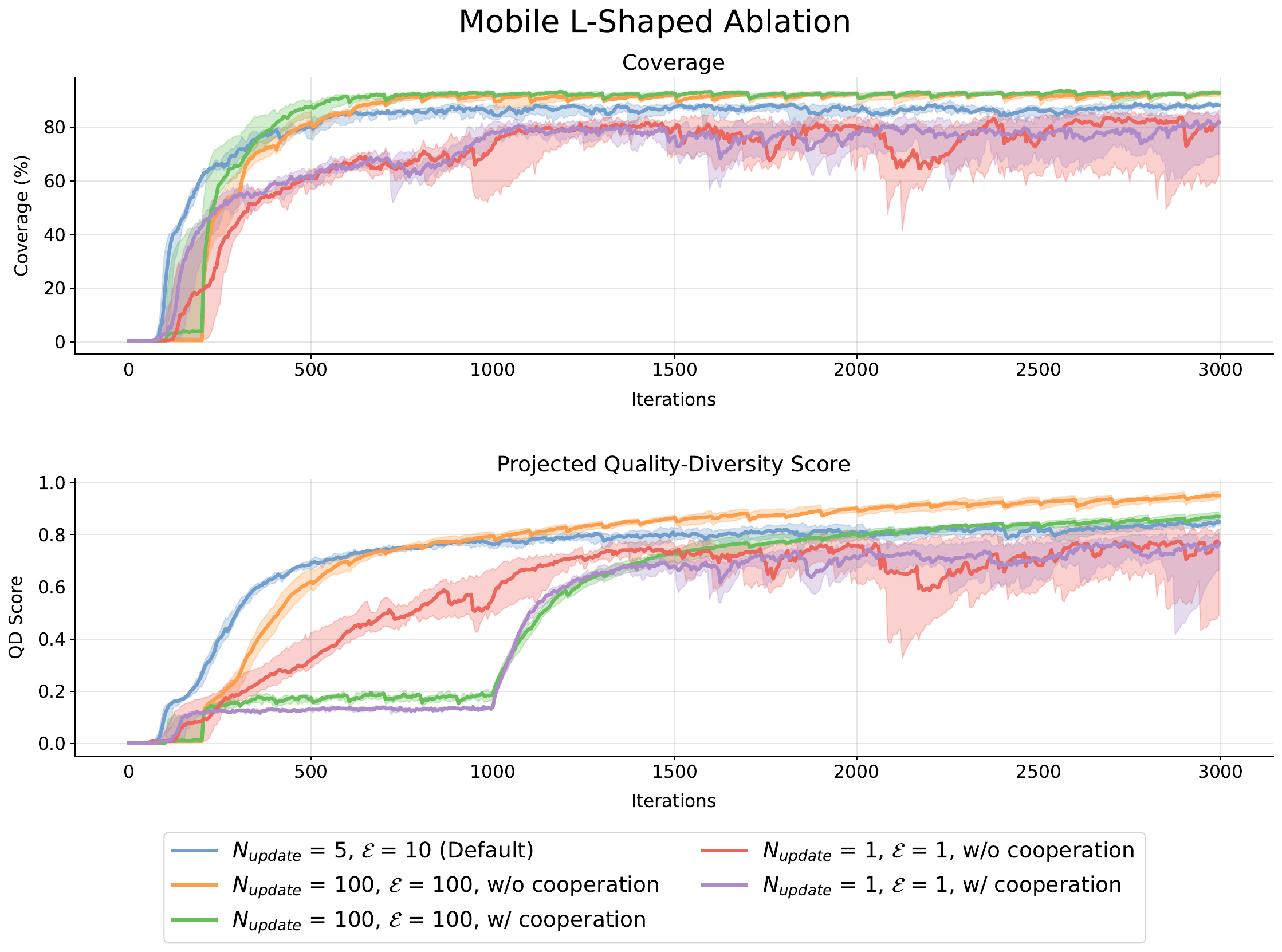}
    \caption{Results of the ablation study regarding the model's update frequency, the number of training epochs, and the use of the cooperation component on the mobile robot maze experiment. Each graph line represents the median and the 25th and 75th quantiles over ten runs for the corresponding algorithm.}
    \label{fig:ablation_all}
\end{figure}

In order to identify whether {\algo} is robust to hyper-parameters we perform some ablation studies. In particular, we want to investigate how the combination of the model's update frequency, the training epochs, and the use of the cooperation component, affect the performance of our algorithm. For these experiments, we used some extreme values for the mentioned hyperparameters to investigate our algorithm's robustness. We perform the experiments in the L-shaped mobile robot environment. The results showcase that when we update the model after every iteration of the evolution phase and train the model for 1 epoch, the performance is suboptimal despite the use of the cooperation component (Fig.~\ref{fig:ablation_all}). This is anticipated since the behavior space grid changes continuously, but also the model fails to converge fast enough due to the low number of gradient updates. Conversely, when we update the model every 100 iterations for 100 epochs, the performance is similar to our original hyperparameters~(answers to Question~3). Although one could argue that this is more selection of hyperparameters compared to the ones we use in our main experiments, we should emphasize that updating the model after every 100 iterations raises memory and wall-time issues. For every iteration, one should store all the raw data produced by the populations, which can easily lead to reaching that maximum memory capacity, especially when working with high-dimensional data like images, while significantly slowing down the model's update.

\begin{figure}[h]
    \centering
    \includegraphics[width=0.9\linewidth]{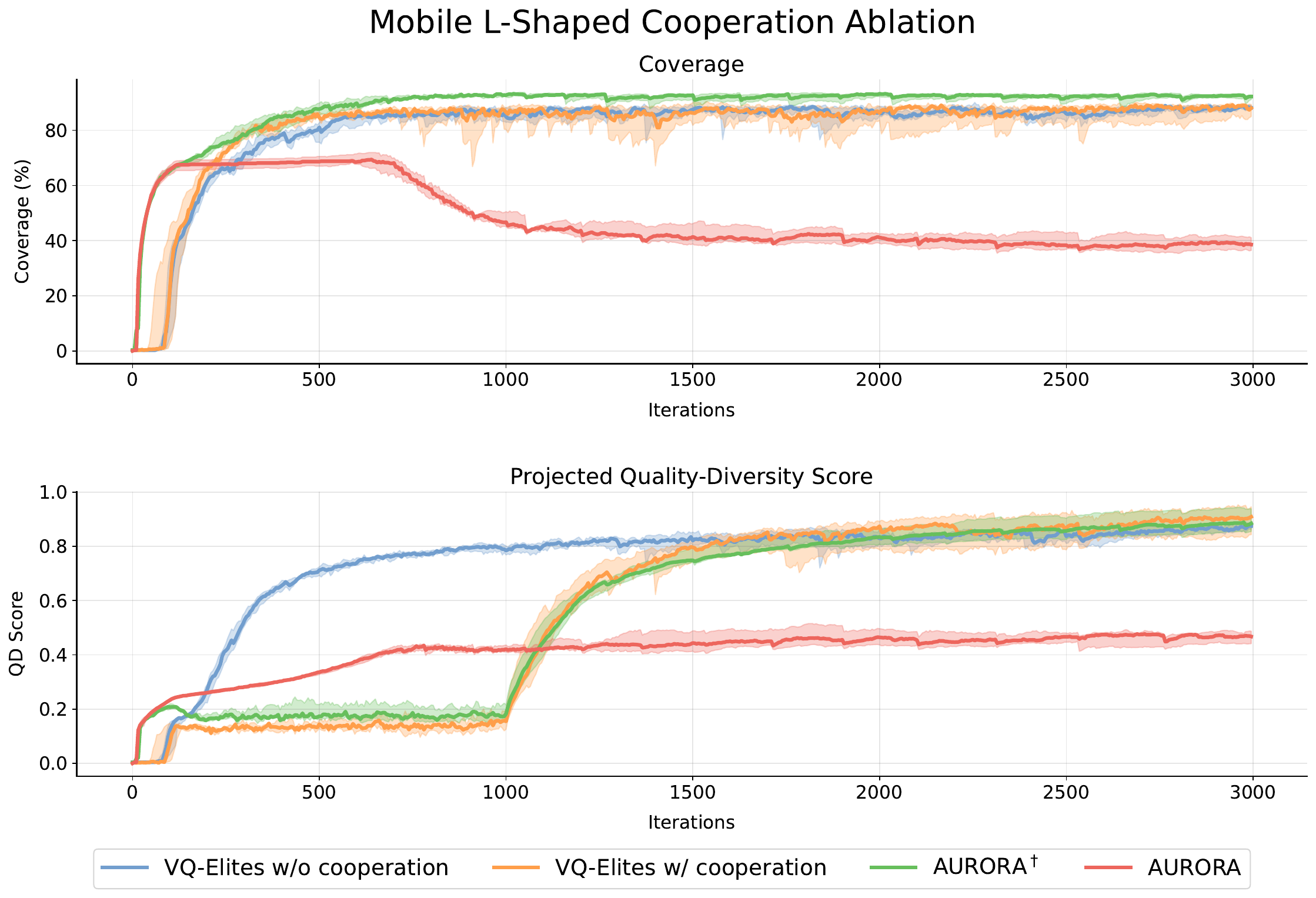}
    \caption{Illustration of the effect of the cooperation component on {\algo} and AURORA.}
    \label{fig:cooperation_abblation}
\end{figure}

Furthermore, we investigate the importance of the cooperation component on our algorithms as well as on the AURORA algorithm. As stated, {\algo} is quite robust that the cooperation does not have any impact on its performance at convergence, since the VQ-VAE model, eventually, converges to an optimal grid structure~(answers to Question~4). As expected, while the cooperation mechanism is active, the PQD score may fluctuate or remain low, since the algorithm intentionally allows lower-quality solutions to enter the archive in favor of promoting diversity. On the other hand, the AURORA algorithm performs poorly without this component (Fig.~\ref{fig:cooperation_abblation}).

\begin{figure}[h]
    \centering
    \includegraphics[width=0.9\linewidth]{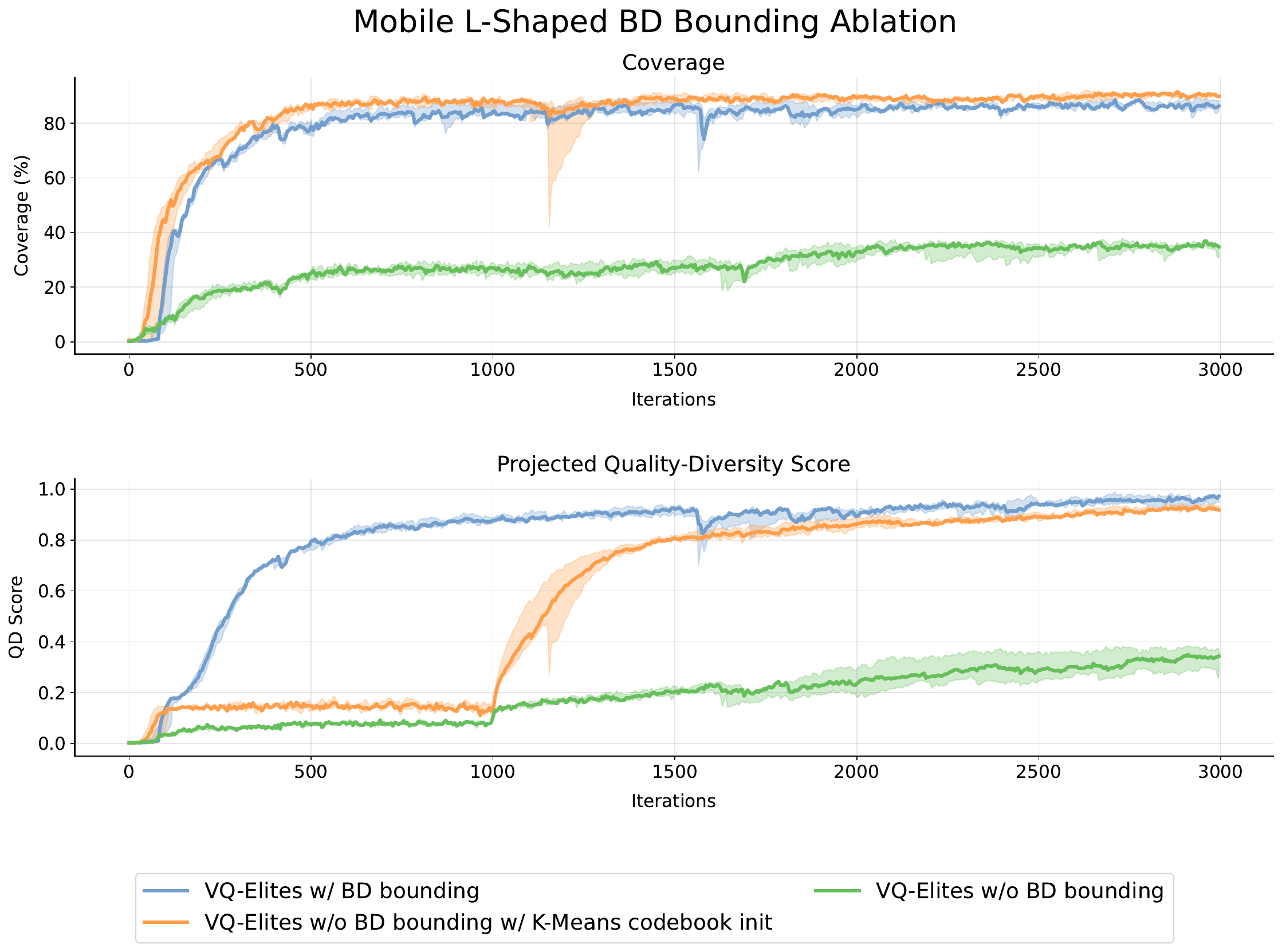}
    \caption{Results of the ablation study regarding the behavior space bounding component. Each graph line represents the median and the 25th and 75th quantiles over ten runs for the corresponding algorithm.}
    \label{fig:ablation_bdb}
\end{figure}

Lastly, we performed an ablation study on the importance of behavior space bound in our framework, and through that, a very interesting result emerged. Essentially, it is evident from the Fig. \ref{fig:ablation_bdb} that without the behavior bounding our approach's performance is hindered significantly. However, we investigated another option, which is to remove behavior space bounding but initialize the codebook of the VQ-VAE to $[-1, 1]$. By doing so, we are able to achieve similar performance with our proposed algorithm. This indicates that providing a structured codebook to the model boosts the convergence of the model substantially, since the centers are well dispersed leading to much stronger commitment loss even in the initial stages of the learning.

\section{Discussion and Future Work}\label{sec:discussion}

While \algo{} demonstrates promising performance across domains, several open questions and limitations merit further reflection. One such issue is the observed degradation in QD performance during phases where the cooperation mechanism is active. Although cooperation facilitates faster exploration and enables model improvement through diverse samples, it can temporarily lower the archive’s overall quality, as lower-performing individuals are admitted to encourage diversity. Balancing this trade-off between diversity promotion and performance degradation remains an open challenge. Nevertheless, \algo{} in general does not need the cooperation mechanism, and thus does not suffer from the above limitation.

Another limitation concerns the scalability of the approach to very high-dimensional behavior spaces, especially when the raw descriptors are unstructured. While \algo{} effectively compresses these spaces using a learned representation, the VQ-VAE model may struggle to disentangle meaningful features in complex environments without careful design. This limitation could potentially be mitigated by incorporating explicit structure into the latent space. For instance, enforcing Riemannian~\cite{lin2008riemannian} or $SO(3)$ manifold constraints may lead to more meaningful and interpretable representations for behaviors with geometric or rotational properties. Such structure could be introduced either through architectural constraints or auxiliary objective functions during training.

Computational requirements are another important consideration. The combined cost of online training, archive updates, and behavior space encoding can be substantial, especially for real-time or hardware-in-the-loop applications. Although \algo{} is compatible with asynchronous evaluation and model updates, future work should explore strategies for reducing model complexity, leveraging distillation, or introducing sparse model updates.

Moreover, while \algo{} was designed to be general and task-agnostic, this generality may come at the cost of suboptimality in highly structured domains. Future research could explore hybrid designs where partial task knowledge or hand-crafted descriptors are used to guide the model in early phases, later transitioning to fully learned representations as more data becomes available. Additionally, techniques such as attention mechanisms~\cite{vaswani2017attention} or multi-modal inputs (e.g., fusing proprioceptive and visual data) could enhance representation quality in challenging domains.

Our results also reveal broader challenges in evaluating and designing unsupervised QD algorithms. Existing metrics, such as coverage and QD-score, provide limited insight into how well the algorithm captures the structure of the unknown behavior space. To address this, we introduced the Effective Diversity Ratio (EDR) and the Coverage Diversity Score (CDS), which offer a more nuanced perspective on solution redundancy and behavioral expressiveness. However, these metrics are still proxies, and future work is needed to develop principled evaluation tools that align with the goals of open-ended exploration and meaningful behavior diversity.

Finally, in the future, we want to incorporate generative modeling techniques such as diffusion~\cite{ho2020denoising,yang2023diffusion}, or flow-matching~\cite{lipman2022flow}, in order to model the distribution of solutions and generate more high-quality and diverse solutions. An ideal outcome would be to use {\algo} for finding the minimum amount of solutions, and then via the generative modeling techniques, we would be able to fill the archive successfully without being limited by the run times of the evolutionary algorithm and its sequential nature. Another interesting direction would be an epistemic uncertainty-guided evolution frame. The intuition here lies in the fact that epistemic uncertainty decreases when the number of data we train our model increases~\cite{hullermeier2021aleatoric}, hence, we can use the epistemic uncertainty of the model as a way to guide the QD evolution into more uncertain behaviors, with the goal of reducing the epistemic uncertainty of the VQ-VAE model, and consequently creating the most optimal behavioral space possible. For this purpose, we can borrow ideas from surrogate-assisted QD algorithms~\cite{kent2024bayesian}. \change{Lastly, we can combine the unsupervised QD with elements of information theory, specifically the normalized entropy \cite{wilcox1967indices}, to devise a more concrete metric to assess the quality of the behavior space.}

\section{Conclusion}\label{sec:conclusion}

In this work, we presented the Vector Quantized-Elites ({\algo}) algorithm, an unsupervised and problem-agnostic variation of the MAP-Elites algorithm. Our algorithm requires no prior knowledge of the task at hand to provide high-quality and diverse solutions. Additionally, we proposed two key components, the behavioral space bounding and cooperation components, which can be integrated into other unsupervised QD algorithms to mitigate the difficulties between the evolution and the update of the unsupervised model. We also introduced two novel metrics, the Effective Diversity Ratio and the Coverage Diversity Score in order to facilitate the evaluation of unsupervised QD algrithms. Our work is deployed on a mobile robot, a robotic arm, and a MiniGrid-type game task, while its performance is evaluated on numerous experiments. In all scenarios, our algorithm performs as well as or better than the AURORA algorithm (even when coupled with our two novel components, {\aur}) in terms of coverage and QD score, yet it outperforms them in the EDR and CDS categories, making it a strong candidate to tackle many optimization problems.

\section*{Acknowledgments}
{\small
This work was supported by the Hellenic Foundation for Research and Innovation (H.F.R.I.) under the ``3rd Call for H.F.R.I. Research Projects to support Post-Doctoral Researchers'' (Project Acronym: NOSALRO, Project Number: 7541).
}

\bibliographystyle{ieeetr}
\bibliography{references}

@inproceedings{edl,
  author    = {V{\'{\i}}ctor Campos and
               Alexander Trott and
               Caiming Xiong and
               Richard Socher and
               Xavier Gir{\'{o}}{-}i{-}Nieto and
               Jordi Torres},
  title     = {Explore, Discover and Learn: Unsupervised Discovery of State-Covering
               Skills},
  booktitle   = {International Conference on Machine Learning (ICML)},
  year      = {2020}
}

@INPROCEEDINGS{agrl,
  author={Tsakonas, Constantinos G. and Chatzilygeroudis, Konstantinos I.},
  booktitle={2023 14th International Conference on Information, Intelligence, Systems \& Applications (IISA)}, 
  title={Effective Skill Learning via Autonomous Goal Representation Learning}, 
  year={2023},
  volume={},
  number={},
  pages={1-8},
  doi={10.1109/IISA59645.2023.10345879}}

@article{chevalier2018minimalistic,
  title={Minimalistic gridworld environment for openai gym},
  author={Chevalier-Boisvert, Maxime and Willems, Lucas and Pal, Suman},
  year={2018}
}

@misc{mouret2015illuminatingsearchspacesmapping,
      title={Illuminating search spaces by mapping elites}, 
      author={Jean-Baptiste Mouret and Jeff Clune},
      year={2015},
      eprint={1504.04909},
      archivePrefix={arXiv},
      primaryClass={cs.AI},
      url={https://arxiv.org/abs/1504.04909}, 
}

@INPROCEEDINGS{loco_srbd,
  author={Chatzilygeroudis, Konstantinos I. and Tsakonas, Constantinos G. and Vrahatis, Michael N.},
  booktitle={2023 14th International Conference on Information, Intelligence, Systems \& Applications (IISA)}, 
  title={Evolving Dynamic Locomotion Policies in Minutes}, 
  year={2023},
  volume={},
  number={},
  pages={1-8},
  doi={10.1109/IISA59645.2023.10345937}}

@article{cully2017quality,
  title={Quality and diversity optimization: A unifying modular framework},
  author={Cully, Antoine and Demiris, Yiannis},
  journal={IEEE Transactions on Evolutionary Computation},
  volume={22},
  number={2},
  pages={245--259},
  year={2017},
  publisher={IEEE}
}

@incollection{chatzilygeroudis2021quality,
  title={Quality-diversity optimization: a novel branch of stochastic optimization},
  author={Chatzilygeroudis, Konstantinos and Cully, Antoine and Vassiliades, Vassilis and Mouret, Jean-Baptiste},
  booktitle={Black Box Optimization, Machine Learning, and No-Free Lunch Theorems},
  pages={109--135},
  year={2021},
  publisher={Springer}
}

@article{vaswani2017attention,
  title={Attention is all you need},
  author={Vaswani, Ashish and Shazeer, Noam and Parmar, Niki and Uszkoreit, Jakob and Jones, Llion and Gomez, Aidan N and Kaiser, {\L}ukasz and Polosukhin, Illia},
  journal={Advances in neural information processing systems},
  volume={30},
  year={2017}
}

@article{lin2008riemannian,
  title={Riemannian manifold learning},
  author={Lin, Tong and Zha, Hongbin},
  journal={IEEE transactions on pattern analysis and machine intelligence},
  volume={30},
  number={5},
  pages={796--809},
  year={2008},
  publisher={IEEE}
}

@article{ho2020denoising,
  title={Denoising diffusion probabilistic models},
  author={Ho, Jonathan and Jain, Ajay and Abbeel, Pieter},
  journal={Advances in neural information processing systems},
  volume={33},
  pages={6840--6851},
  year={2020}
}

@article{yang2023diffusion,
  title={Diffusion models: A comprehensive survey of methods and applications},
  author={Yang, Ling and Zhang, Zhilong and Song, Yang and Hong, Shenda and Xu, Runsheng and Zhao, Yue and Zhang, Wentao and Cui, Bin and Yang, Ming-Hsuan},
  journal={ACM Computing Surveys},
  volume={56},
  number={4},
  pages={1--39},
  year={2023},
  publisher={ACM New York, NY, USA}
}

@article{lipman2022flow,
  title={Flow matching for generative modeling},
  author={Lipman, Yaron and Chen, Ricky TQ and Ben-Hamu, Heli and Nickel, Maximilian and Le, Matt},
  journal={arXiv preprint arXiv:2210.02747},
  year={2022}
}

@article{kent2024bayesian,
  title={Bayesian optimisation for quality diversity search with coupled descriptor functions},
  author={Kent, Paul and Gaier, Adam and Mouret, Jean-Baptiste and Branke, Juergen},
  journal={IEEE Transactions on Evolutionary Computation},
  year={2024},
  publisher={IEEE}
}

@article{hullermeier2021aleatoric,
  title={Aleatoric and epistemic uncertainty in machine learning: An introduction to concepts and methods},
  author={H{\"u}llermeier, Eyke and Waegeman, Willem},
  journal={Machine learning},
  volume={110},
  number={3},
  pages={457--506},
  year={2021},
  publisher={Springer}
}

@article{paolo2024discovering,
  title={Discovering and exploiting sparse rewards in a learned behavior space},
  author={Paolo, Giuseppe and Coninx, Miranda and Laflaqui{\`e}re, Alban and Doncieux, Stephane},
  journal={Evolutionary Computation},
  volume={32},
  number={3},
  pages={275--305},
  year={2024},
  publisher={MIT Press 255 Main Street, 9th Floor, Cambridge, Massachusetts 02142, USA~…}
}

@inproceedings{nadizar2024searching,
  title={Searching for a Diversity of Interpretable Graph Control Policies},
  author={Nadizar, Giorgia and Medvet, Eric and Wilson, Dennis},
  booktitle={Proceedings of the Genetic and Evolutionary Computation Conference},
  pages={933--941},
  year={2024}
}

@inproceedings{lancucki2020robust,
  title={Robust training of vector quantized bottleneck models},
  author={{\L}a{\'n}cucki, Adrian and Chorowski, Jan and Sanchez, Guillaume and Marxer, Ricard and Chen, Nanxin and Dolfing, Hans JGA and Khurana, Sameer and Alum{\"a}e, Tanel and Laurent, Antoine},
  booktitle={2020 International Joint Conference on Neural Networks (IJCNN)},
  pages={1--7},
  year={2020},
  organization={IEEE}
}

@ARTICLE{map-elites-voronoi,
  author={Vassiliades, Vassilis and Chatzilygeroudis, Konstantinos and Mouret, Jean-Baptiste},
  journal={IEEE Transactions on Evolutionary Computation}, 
  title={Using Centroidal Voronoi Tessellations to Scale Up the Multidimensional Archive of Phenotypic Elites Algorithm}, 
  year={2018},
  volume={22},
  number={4},
  pages={623-630},
  doi={10.1109/TEVC.2017.2735550}}

@ARTICLE{aurora,
  author={Grillotti, Luca and Cully, Antoine},
  journal={IEEE Transactions on Evolutionary Computation}, 
  title={{Unsupervised Behavior Discovery With Quality-Diversity Optimization}}, 
  year={2022},
  volume={26},
  number={6},
  pages={1539-1552},
  doi={10.1109/TEVC.2022.3159855}}

@inproceedings{cully2019autonomous,
  title={Autonomous skill discovery with quality-diversity and unsupervised descriptors},
  author={Cully, Antoine},
  booktitle={Proceedings of the Genetic and Evolutionary Computation Conference},
  pages={81--89},
  year={2019}
}

@inproceedings{vqvae_paper,
author = {van den Oord, Aaron and Vinyals, Oriol and Kavukcuoglu, Koray},
title = {Neural discrete representation learning},
year = {2017},
isbn = {9781510860964},
publisher = {Curran Associates Inc.},
address = {Red Hook, NY, USA},
booktitle = {Proceedings of the 31st International Conference on Neural Information Processing Systems},
pages = {6309–6318},
numpages = {10},
location = {Long Beach, California, USA},
series = {NIPS'17}
}

@inproceedings{cully2018hierarchical,
  title={Hierarchical behavioral repertoires with unsupervised descriptors},
  author={Cully, Antoine and Demiris, Yiannis},
  booktitle={Proceedings of the Genetic and Evolutionary Computation Conference},
  pages={69--76},
  year={2018}
}

@article{chatzilygeroudis2018reset,
  title={Reset-free trial-and-error learning for robot damage recovery},
  author={Chatzilygeroudis, Konstantinos and Vassiliades, Vassilis and Mouret, Jean-Baptiste},
  journal={Robotics and Autonomous Systems},
  volume={100},
  pages={236--250},
  year={2018},
  publisher={Elsevier}
}

@article{cully2015robots,
  title={Robots that can adapt like animals},
  author={Cully, Antoine and Clune, Jeff and Tarapore, Danesh and Mouret, Jean-Baptiste},
  journal={Nature},
  volume={521},
  number={7553},
  pages={503--507},
  year={2015},
  publisher={Nature Publishing Group}
}

@inproceedings{nilsson2021policy,
  title={Policy gradient assisted map-elites},
  author={Nilsson, Per and Cully, Antoine},
  booktitle={Proceedings of the Genetic and Evolutionary Computation Conference},
  pages={866--875},
  year={2021}
}

@article{pierrot2022qd,
  title={QD-PG: A QD algorithm with policy gradients for sample-efficient exploration},
  author={Pierrot, Tim and Cully, Antoine and others},
  journal={arXiv preprint arXiv:2204.12868},
  year={2022}
}

@inproceedings{ecoffet2021go,
  title={Go-explore: Towards an optimistic initial exploration},
  author={Ecoffet, Adrien and Huizinga, Joost and Lehman, Joel and Stanley, Kenneth O and Clune, Jeff},
  booktitle={International Conference on Machine Learning},
  pages={1517--1528},
  year={2021}
}

@inproceedings{gupta2018unsupervised,
  title={Unsupervised learning for robotic skill discovery},
  author={Gupta, Abhishek and Eysenbach, Benjamin and Finn, Chelsea and Levine, Sergey},
  booktitle={Conference on Robot Learning},
  pages={403--414},
  year={2018}
}

@article{barto2003recent,
  title={Recent advances in hierarchical reinforcement learning},
  author={Barto, Andrew G and Mahadevan, Sridhar},
  journal={Discrete event dynamic systems},
  volume={13},
  number={4},
  pages={341--379},
  year={2003}
}

@inproceedings{konidaris2009skill,
  title={Skill discovery in continuous reinforcement learning domains using skill chaining},
  author={Konidaris, George and Barto, Andrew},
  booktitle={Advances in Neural Information Processing Systems},
  pages={1015--1023},
  year={2009}
}

@article{eysenbach2018diversity,
  title={Diversity is all you need: Learning skills without a reward function},
  author={Eysenbach, Benjamin and Gupta, Abhishek and Ibarz, Julian and Levine, Sergey},
  journal={arXiv preprint arXiv:1802.06070},
  year={2018}
}

@article{gregor2016variational,
  title={Variational intrinsic control},
  author={Gregor, Karol and Rezende, Danilo Jimenez and Danihelka, Ivo and Wierstra, Daan},
  journal={arXiv preprint arXiv:1611.07507},
  year={2016}
}

@inproceedings{vq-vae-2,
 author = {Razavi, Ali and van den Oord, Aaron and Vinyals, Oriol},
 booktitle = {Advances in Neural Information Processing Systems},
 editor = {H. Wallach and H. Larochelle and A. Beygelzimer and F. d\textquotesingle Alch\'{e}-Buc and E. Fox and R. Garnett},
 pages = {},
 publisher = {Curran Associates, Inc.},
 title = {Generating Diverse High-Fidelity Images with VQ-VAE-2},
 url = {https://proceedings.neurips.cc/paper_files/paper/2019/file/5f8e2fa1718d1bbcadf1cd9c7a54fb8c-Paper.pdf},
 volume = {32},
 year = {2019}
}

@misc{walker2021predictingvideovqvae,
      title={Predicting Video with VQVAE}, 
      author={Jacob Walker and Ali Razavi and Aäron van den Oord},
      year={2021},
      eprint={2103.01950},
      archivePrefix={arXiv},
      primaryClass={cs.CV},
      url={https://arxiv.org/abs/2103.01950}, 
}

@article{DBLP:journals/corr/abs-2312-11532,
  author       = {Youngjoon Yoo and
                  Jongwon Choi},
  title        = {Topic-VQ-VAE: Leveraging Latent Codebooks for Flexible Topic-Guided
                  Document Generation},
  journal      = {CoRR},
  volume       = {abs/2312.11532},
  year         = {2023},
  url          = {https://doi.org/10.48550/arXiv.2312.11532},
  doi          = {10.48550/ARXIV.2312.11532},
  eprinttype    = {arXiv},
  eprint       = {2312.11532},
  bibsource    = {dblp computer science bibliography, https://dblp.org}
}

@inproceedings{DBLP:conf/icassp/ZhouBA21,
  author       = {Henry Zhou and
                  Alexei Baevski and
                  Michael Auli},
  title        = {A Comparison of Discrete Latent Variable Models for Speech Representation
                  Learning},
  booktitle    = {{IEEE} International Conference on Acoustics, Speech and Signal Processing,
                  {ICASSP} 2021, Toronto, ON, Canada, June 6-11, 2021},
  pages        = {3050--3054},
  publisher    = {{IEEE}},
  year         = {2021},
  url          = {https://doi.org/10.1109/ICASSP39728.2021.9413680},
  doi          = {10.1109/ICASSP39728.2021.9413680},
  bibsource    = {dblp computer science bibliography, https://dblp.org}
}

@inproceedings{DBLP:conf/icassp/WilliamsZCY21,
  author       = {Jennifer Williams and
                  Yi Zhao and
                  Erica Cooper and
                  Junichi Yamagishi},
  title        = {Learning Disentangled Phone and Speaker Representations in a Semi-Supervised
                  {VQ-VAE} Paradigm},
  booktitle    = {{IEEE} International Conference on Acoustics, Speech and Signal Processing,
                  {ICASSP} 2021, Toronto, ON, Canada, June 6-11, 2021},
  pages        = {7053--7057},
  publisher    = {{IEEE}},
  year         = {2021},
  url          = {https://doi.org/10.1109/ICASSP39728.2021.9413543},
  doi          = {10.1109/ICASSP39728.2021.9413543},
  bibsource    = {dblp computer science bibliography, https://dblp.org}
}

@INPROCEEDINGS{9196819,
  author={Paolo, Giuseppe and Laflaquière, Alban and Coninx, Alexandre and Doncieux, Stephane},
  booktitle={2020 IEEE International Conference on Robotics and Automation (ICRA)}, 
  title={Unsupervised Learning and Exploration of Reachable Outcome Space}, 
  year={2020},
  volume={},
  number={},
  pages={2379-2385},
  doi={10.1109/ICRA40945.2020.9196819}}

@ARTICLE{9716036,
  author={Bossens, David M. and Tarapore, Danesh},
  journal={IEEE Transactions on Evolutionary Computation}, 
  title={Quality-Diversity Meta-Evolution: Customizing Behavior Spaces to a Meta-Objective}, 
  year={2022},
  volume={26},
  number={5},
  pages={1171-1181},
  doi={10.1109/TEVC.2022.3152384}}

@inproceedings{rainbow_teaming,
 author = {Samvelyan, Mikayel and Raparthy, Sharath Chandra and Lupu, Andrei and Hambro, Eric and Markosyan, Aram H. and Bhatt, Manish and Mao, Yuning and Jiang, Minqi and Parker-Holder, Jack and Foerster, Jakob and Rockt\"{a}schel, Tim and Raileanu, Roberta},
 booktitle = {Advances in Neural Information Processing Systems},
 editor = {A. Globerson and L. Mackey and D. Belgrave and A. Fan and U. Paquet and J. Tomczak and C. Zhang},
 pages = {69747--69786},
 publisher = {Curran Associates, Inc.},
 title = {Rainbow Teaming: Open-Ended Generation of Diverse Adversarial Prompts},
 url = {https://proceedings.neurips.cc/paper_files/paper/2024/file/8147a43d030b43a01020774ae1d3e3bb-Paper-Conference.pdf},
 volume = {37},
 year = {2024}
}

@inproceedings{gavel,
 author = {Todd, Graham and Padula, Alexander G. and Stephenson, Matthew and Piette, \'{E}ric and Soemers, Dennis J.N.J. and Togelius, Julian},
 booktitle = {Advances in Neural Information Processing Systems},
 editor = {A. Globerson and L. Mackey and D. Belgrave and A. Fan and U. Paquet and J. Tomczak and C. Zhang},
 pages = {110723--110745},
 publisher = {Curran Associates, Inc.},
 title = {GAVEL: Generating Games via Evolution and Language Models},
 url = {https://proceedings.neurips.cc/paper_files/paper/2024/file/c7b04e4e13bb77996d3ae2ff667231ac-Paper-Conference.pdf},
 volume = {37},
 year = {2024}
}

@inproceedings{me_image_synthesis,
author = {Gaier, Adam and Asteroth, Alexander and Mouret, Jean-Baptiste},
title = {Are quality diversity algorithms better at generating stepping stones than objective-based search?},
year = {2019},
isbn = {9781450367486},
publisher = {Association for Computing Machinery},
address = {New York, NY, USA},
url = {https://doi.org/10.1145/3319619.3321897},
doi = {10.1145/3319619.3321897},
booktitle = {Proceedings of the Genetic and Evolutionary Computation Conference Companion},
pages = {115–116},
numpages = {2},
location = {Prague, Czech Republic},
series = {GECCO '19}
}

@article{pugh2016quality,
  title={Quality diversity: A new frontier for evolutionary computation},
  author={Pugh, Justin K and Soros, Lisa B and Stanley, Kenneth O},
  journal={Frontiers in Robotics and AI},
  volume={3},
  pages={40},
  year={2016},
  publisher={Frontiers Media SA}
}

@techreport{wilcox1967indices,
  title={Indices of Qualitative Variation.},
  author={Wilcox, Allen R},
  year={1967},
  institution={Oak Ridge National Lab.(ORNL), Oak Ridge, TN (United States)}
}




\onecolumn
\appendix
\subsection{Visualization Learned Centers in the MiniGrid Task}

\begin{figure*}[tbh!]
    \centering
    \begin{subfigure}[b]{0.85\linewidth}
      \centering
      \includegraphics[width=0.8\linewidth]{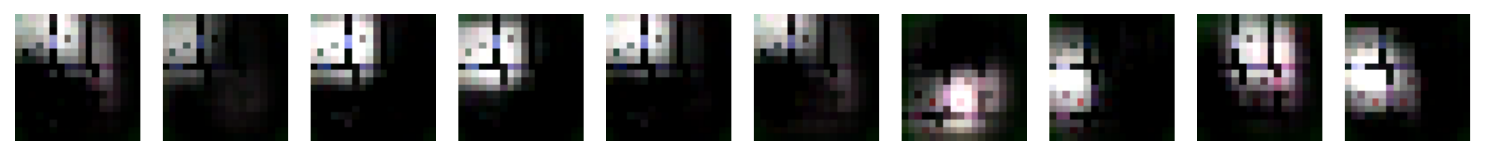}
      \caption{Iteration 40}
    \end{subfigure}%
    \\
    \begin{subfigure}[b]{0.85\linewidth}
      \centering
      \includegraphics[width=0.8\linewidth]{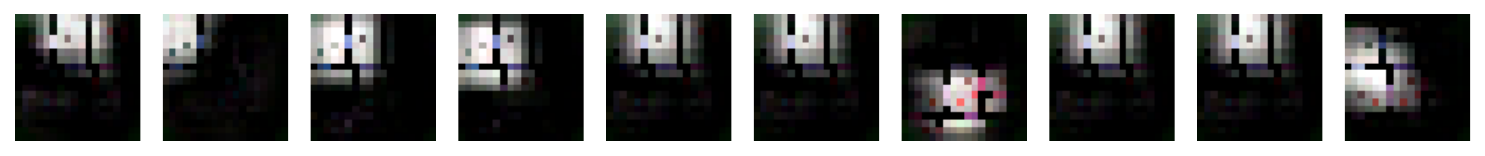}
      \caption{Iteration 60}
    \end{subfigure}%
    \\
    \begin{subfigure}[b]{0.85\linewidth}
      \centering
      \includegraphics[width=0.8\linewidth]{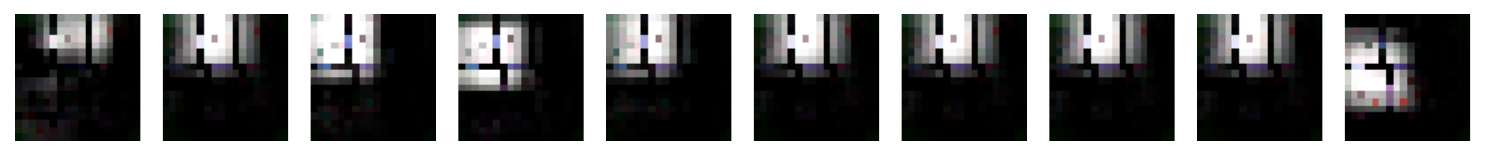}
      \caption{Iteration 80}
    \end{subfigure}%
    \\
    \begin{subfigure}[b]{0.85\linewidth}
      \centering
      \includegraphics[width=0.8\linewidth]{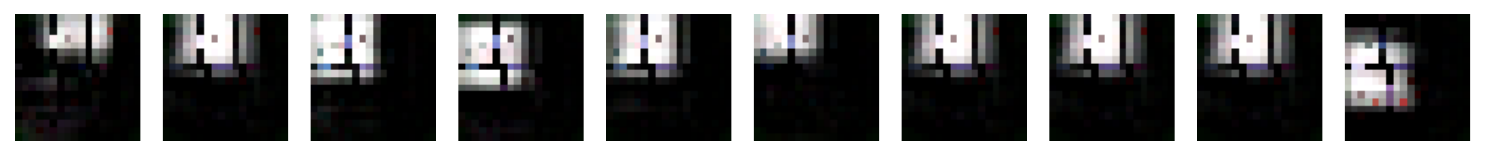}
      \caption{Iteration 100}
    \end{subfigure}%
    \\
    \begin{subfigure}[b]{0.85\linewidth}
      \centering
      \includegraphics[width=0.8\linewidth]{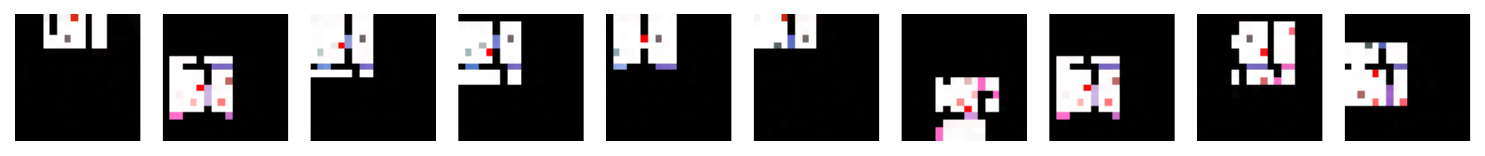}
      \caption{Iteration 2500}
    \end{subfigure}%
    \\
    \begin{subfigure}[b]{0.85\linewidth}
      \centering
      \includegraphics[width=0.8\linewidth]{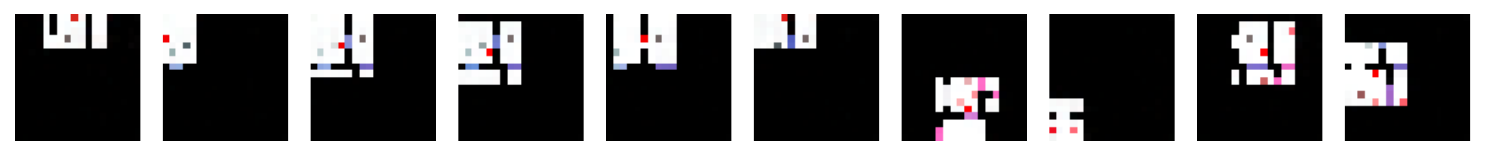}
      \caption{Iteration 5000}
    \end{subfigure}%
    \\
    \begin{subfigure}[b]{0.85\linewidth}
      \centering
      \includegraphics[width=0.8\linewidth]{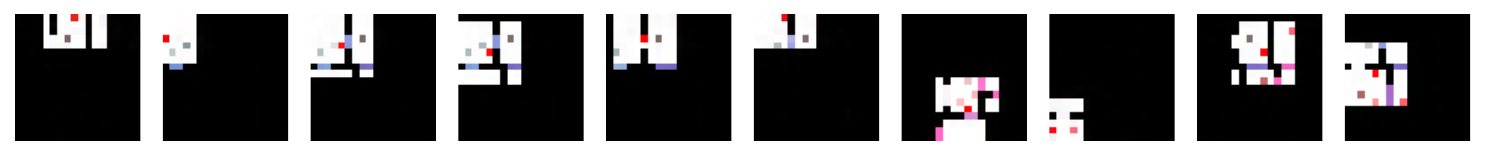}
      \caption{Iteration 7500}
    \end{subfigure}%
    \\
    \begin{subfigure}[b]{0.85\linewidth}
      \centering
      \includegraphics[width=0.8\linewidth]{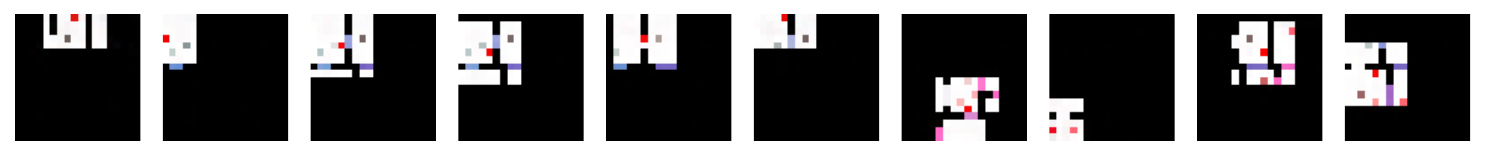}
      \caption{Iteration 10000}
    \end{subfigure}%
    \caption{Visualization of the reconstructed learned centers over the course of the {\algo} execution.}
    \label{fig:centers_rec}
\end{figure*}

Fig~\ref{fig:centers_rec} showcases the reconstructed learned centers of {\algo} over the course of the evolution process. Specifically, we include images from the initial iterations 40, 60, 80, and 100, where the VQ-VAE model has not fully converged yet. Then, we include the reconstructions for iterations 2500, 5000, 7500, and 10000, where the model has started converging and there are only some final refinements in the centers, as evident from the reconstructions.

\subsection{Unsupervised Models and Policies}
\subsubsection{Mobile Experiments}

The mobile robot's experiment policy and encoder-decoder architecture are described in Tables \ref{tab:mobile_policy} and \ref{tab:mobile_model}.

\begin{center}
\begin{table}[!h]
\setlength{\tabcolsep}{2cm}
\centering
\begin{tabular}{l l} 
\toprule
\textbf{Layer} & \textbf{Output Size} \\ \midrule
Input   & [4] \\
Linear   & [16] \\
ReLU     & [16] \\
Linear   & [2] \\
Tanh     & [2] \\ \bottomrule
\end{tabular}
\caption{Mobile experiment policy architecture details.}
\label{tab:mobile_policy}
\end{table}
\end{center}

\begin{center}
\begin{table}[h!]
\renewcommand{\arraystretch}{0.8}
\setlength{\tabcolsep}{1.0cm}
\centering
\begin{tabular}{l l c c}
\toprule
\textbf{Layer} & \textbf{Output Size} & \textbf{Kernel Size} & \textbf{Stride} \\
\midrule
\multicolumn{4}{c}{\textbf{Encoder}} \\
\midrule
Input     & [1,\,64,\,64] & -- & -- \\
Conv2d     & [64,\,61,\,61] & 4 & 1 \\
BatchNorm2d & [64,\,61,\,61] & -- & -- \\
GELU        & [64,\,61,\,61] & -- & -- \\
MaxPool2d   & [64,\,29,\,29] & 4 & 2 \\
Dropout     & [64,\,29,\,29] & -- & -- \\
Conv2d      & [32,\,26,\,26] & 4 & 1 \\
BatchNorm2d & [32,\,26,\,26] & -- & -- \\
GELU        & [32,\,26,\,26] & -- & -- \\
MaxPool2d   & [32,\,12,\,12] & 4 & 2 \\
Flatten   & [4608] & -- & -- \\
Linear   & [256] & -- & -- \\
GELU     & [256] & -- & -- \\
Linear   & [128] & -- & -- \\
GELU     & [128] & -- & -- \\
Linear   & [64]  & -- & -- \\
GELU     & [64]  & -- & -- \\
Linear   & [2]   & -- & -- \\
Tanh & [2]   & -- & -- \\
\midrule
\multicolumn{4}{c}{\textbf{Decoder}} \\
\midrule
Linear & [64] & -- & -- \\
GELU   & [64] & -- & -- \\
Linear & [128] & -- & -- \\
GELU   & [128] & -- & -- \\
Linear & [256] & -- & -- \\
GELU   & [256] & -- & -- \\
Linear & [4608] & -- & -- \\
GELU   & [4608] & -- & -- \\
Unflatten       & [32,\,12,\,12] & -- & -- \\
ConvTranspose2d & [64,\,15,\,15] & 4 & 1 \\
BatchNorm2d     & [64,\,15,\,15] & -- & -- \\
GELU            & [64,\,15,\,15] & -- & -- \\
Dropout         & [64,\,15,\,15] & -- & -- \\
ConvTranspose2d & [32,\,31,\,31] & 3 & 2 \\
BatchNorm2d     & [32,\,31,\,31] & -- & -- \\
GELU            & [32,\,31,\,31] & -- & -- \\
Dropout         & [32,\,31,\,31] & -- & -- \\
ConvTranspose2d & [1,\,64,\,64]  & 3 & 2 \\
Sigmoid      & [1,\,64,\,64]  & -- & -- \\ \bottomrule
\end{tabular}
\caption{Mobile experiment Encoder-Decoder architecture details.}
\label{tab:mobile_model}
\end{table}
\end{center}

\subsubsection{Robotic Arm Experiments}

In Table~\ref{tab:arm_policy} and Table~\ref{tab:arm_model}, we describe the robotic arm's policy and encoder-decoder architecture. We should note that for the policy, we utilize the Gaussian activation function, which helps the policy to produce smoother outputs, and it is defined as follows: $f(x) = \exp(-x^2)$.
\begin{center}
\begin{table}[h!]
\setlength{\tabcolsep}{2cm}
\centering
\begin{tabular}{c c}
\toprule
\textbf{Layer} & \textbf{Output Size} \\ \midrule
Input   & [6] \\
Linear   & [32] \\
Gaussian     & [32] \\
Linear   & [32] \\
Gaussian     & [32] \\
Linear   & [6] \\ \bottomrule
\end{tabular}
\caption{Robotic Arm experiment policy architecture details.}
\label{tab:arm_policy}
\end{table}
\end{center}
\begin{center}
\setlength{\tabcolsep}{2cm}
\begin{table}[h!]
\centering
\begin{tabular}{c c}
\toprule
\textbf{Layer} & \textbf{Output Size} \\
\midrule
\multicolumn{2}{c}{\textbf{Encoder}} \\
\midrule
Input & [6] \\
Linear   & [64] \\
ReLU     & [64] \\
Linear   & [64]\\
ReLU     & [64]\\
Linear   & [5] \\
Tanh & [5]  \\
\midrule
\multicolumn{2}{c}{\textbf{Decoder}} \\
\midrule
Linear & [64] \\
ReLU   & [64] \\
Linear & [64] \\
ReLU   & [64] \\
Linear & [6] \\
\bottomrule
\end{tabular}
\caption{Robotic Arm experiment Encoder-Decoder architecture details.}
\label{tab:arm_model}
\end{table}
\end{center}
\subsubsection{MiniGrid Experiment}
At last, the policy and model architectures for the MiniGrid experiment are shown at Tables~\ref{tab:minigrid_policy} and~\ref{tab:minigrid_model}, respectively.
\begin{center}
\begin{table}[h!]
\setlength{\tabcolsep}{1cm}
\centering
\begin{tabular}{c c c c}
\toprule
\textbf{Layer} & \textbf{Output Size} & \textbf{Kernel Size} & \textbf{Stride} \\ \midrule
Input     & [3,\,18,\,18] & -- & -- \\
Conv2d     & [8,\,9,\,9] & 3 & 2 \\
SiLU        & [8,\,9,\,9] & -- & -- \\
Conv2d      & [16,\,5,\,5] & 3 & 2 \\
SiLU        & [16,\,5,\,5] & -- & -- \\
Flatten        & [400] & -- & -- \\
Linear   & [16] & -- & -- \\
SiLU     & [16]  & -- & -- \\
Linear   & [5]   & -- & -- \\
Softmax & [5]   & -- & -- \\
\bottomrule
\end{tabular}
\caption{MiniGrid policy network.}
\label{tab:minigrid_policy}
\end{table}
\end{center}
\begin{center}
\begin{table}[h!]
\setlength{\tabcolsep}{1cm}
\centering
\begin{tabular}{l l c c}
\toprule
\textbf{Layer} & \textbf{Output Size} & \textbf{Kernel Size} & \textbf{Stride} \\ \midrule
\multicolumn{4}{c}{\textbf{Encoder}} \\ \midrule
Input       & [1,\,18,\,18] & -- & -- \\
Conv2d      & [16,\,16,\,16] & 3 & 1 \\
SiLU        & [16,\,16,\,16] & -- & -- \\
Dropout     & [16,\,16,\,16] & -- & -- \\
Conv2d      & [32,\,14,\,14] & 3 & 1 \\
SiLU        & [32,\,14,\,14] & -- & -- \\
Dropout     & [32,\,14,\,14] & -- & -- \\
Conv2d      & [64,\,12,\,12] & 3 & 1 \\
SiLU        & [64,\,12,\,12] & -- & -- \\
Flatten     & [9216] & -- & -- \\
Linear   & [256] & -- & -- \\
SilLU    & [256] & -- & -- \\
Linear   & [128] & -- & -- \\
SiLU     & [128] & -- & -- \\
Linear   & [5]   & -- & -- \\
Tanh     & [5]   & -- & -- \\ \midrule
\multicolumn{4}{c}{\textbf{Decoder}} \\ \midrule
Linear & [128] & -- & -- \\
SiLU   & [128] & -- & -- \\
Linear & [256] & -- & -- \\
SiLU   & [256] & -- & -- \\
Linear & [9216] & -- & -- \\
SiLU   & [9216] & -- & -- \\
Unflatten       & [64,\,12,\,12] & -- & -- \\
ConvTranspose2d & [64,\,14,\,14] & 4 & 1 \\
SiLU            & [64,\,14,\,14] & -- & -- \\
Dropout         & [64,\,14,\,14] & -- & -- \\
ConvTranspose2d & [32,\,16,\,16] & 3 & 2 \\
GELU            & [32,\,16,\,16] & -- & -- \\
Dropout         & [32,\,16,\,16] & -- & -- \\
ConvTranspose2d & [3,\,18,\,18]  & 3 & 2 \\
\bottomrule
\end{tabular}
\caption{MiniGrid experiment Encoder-Decoder architecture details.}
\label{tab:minigrid_model}
\end{table}
\end{center}
\subsection{AURORA-Based Experiments Hyperparameters}
In Table~\ref{tab:hyperparameters_aur}, we report the hyperparameters used for the two AURORA variants, the {\aur} and the vanilla AURORA. We used exactly the same hyper-parameters as reported in the original paper (and code), except for the maximum archive size (\emph{Archive Size} in the table), which we set to a smaller value due to memory constraints of our setup, and the target archive size, which we set to values relevant for our experiments.
\begin{table}[h]
 \renewcommand{\arraystretch}{0.5}
 \setlength{\tabcolsep}{0.9cm}
 \caption{List of all the hyperparameters used for each type of environment}
 \label{tab:hyperparameters_aur}
 \centering
\begin{tabular}{ c c c c }
 \toprule
 \multicolumn{4}{c}{Experiment Hyperparameters} \\
 \midrule
 Experiment & Mobile robot    & Robotic arm  & MiniGrid \\
 \midrule
 \multicolumn{4}{c}{Common} \\
 \midrule
 Iterations   & $3 \times 10^3$    & $3 \times 10^3$ & $10^4$ \\
 Population size & $128$ & $128$ & $128$ \\
 $|\mathcal{C}|$ \scriptsize(target archive size) &   $2000$  & $1500$ & $400$ \\
 Archive Size &   $5000$  & $2500$ & $800$ \\
 Raw BD size &   $[1, 64, 64]$ & $6$ & $[3, 18, 18]$  \\
 BD/Latent size &   $2$  & $5$ & $5$ \\
 $\mathcal{E}$ &   $100$  & $100$ & $100$ \\
 $lr$ &   $7\times10^{-4}$  & $7\times10^{-4}$  & $7\times10^{-4}$ \\
 Batch size &   $64$  & $64$  & $64$ \\
 \midrule
 \multicolumn{4}{c}{\aur} \\
 \midrule
 Init Min Distance & $1\times10^{-5}$ & $1\times10^{-5}$ & $1\times10^{-5}$ \\
 Min Distance & $1\times10^{-5}$ & $1\times10^{-5}$ & $1\times10^{-5}$ \\
 Max Distance & $1$ & $1$ & $1$ \\
  $K_\text{CSC}$ & $5\times10^{-4}$ & $5\times10^{-4}$ & $5\times10^{-4}$ \\
 \midrule
 \multicolumn{4}{c}{AURORA} \\
 \midrule
 Init Min Distance & $1\times10^{-5}$ & $1\times10^{-5}$ & $1\times10^{-5}$ \\
 Min Distance & $1\times10^{-5}$ & $1\times10^{-5}$ & $1\times10^{-5}$ \\
 Max Distance & $1\times10^{5}$ & $1\times10^{5}$ & $1\times10^{5}$ \\
  $K_\text{CSC}$ & $5\times10^{-6}$ & $5\times10^{-6}$ & $5\times10^{-6}$ \\
 \bottomrule
\end{tabular}
\end{table}
\subsection{Statistical Significance Tests}
In Table~\ref{tab:pvalues}, we report the $p$-values of the null-hypothesis significance test of the coverage at the final iteration between the algorithms across all the experiments. Overall, the outcome of this analysis says that we can trust the conclusions drawn from the experiments and the figures.
\begin{table}[!h]
\centering
\caption{$p$-value of the Coverage Across Tasks}
\begin{tabular}{ll|ccccc}
\toprule \\
 Algorithm 1 & Algorithm 2 & \makecell{Mobile\\Obstacle-Free} & 
\makecell{Mobile\\L-Shaped} & 
\makecell{Arm Pose\\Reaching} & 
\makecell{Arm Pose\\Reaching\\w/ Joint Constraints} & 
\makecell{MiniGrid\\Exploration} \\
\midrule
\algo   & MAP-Elites & 1.1e-05 & 0.007   & 0.0079 & 0.0079 & 0.02 \\
\algo   & \aur       & 4.3e-05 & 0.22 & 0.0079 & 0.0079 & 0.0079 \\
\algo   & AURORA     & 1.1e-05 & 0.007 & 0.0079 & 0.0079 & 0.42 \\
\aur    & MAP-Elites & 1.1e-05 & 0.007 & 0.42   & 0.42   & 0.10 \\
\aur    & AURORA     & 1.1e-05 & 0.007 & 0.15   & 0.15   & 0.0079 \\
AURORA  & MAP-Elites & 1.1e-05 & 0.22   & 0.42   & 0.42   & 0.02 \\
\end{tabular}
\end{table}
\label{tab:pvalues}

\end{document}